\documentclass[sigconf]{acmart}
\usepackage{booktabs}
\usepackage{makecell}
\usepackage{newunicodechar}
\usepackage[normalem]{ulem}  

\usepackage[utf8]{inputenc}
\DeclareUnicodeCharacter{221A}{\checkmark}  
\DeclareUnicodeCharacter{00D7}{$\times$}     
\newcommand{\longdash}{\rule[0.5ex]{3em}{0.4pt}} 

\AtBeginDocument{%
  }

\setcopyright{acmlicensed}
\copyrightyear{2018}
\acmYear{2018}
\acmDOI{XXXXXXX.XXXXXXX}
\acmConference[Conference acronym 'XX]{Make sure to enter the correct
  conference title from your rights confirmation email}{June 03--05,
  2018}{Woodstock, NY}
\acmISBN{978-1-4503-XXXX-X/2018/06}

\begin{document}

\title{SurgWound-Bench: A Benchmark for Surgical Wound Diagnosis}


\author{Jiahao Xu}
\affiliation{%
  \institution{The Ohio State University}
  \city{Columbus}
  \state{Ohio}
  \country{USA}
  }
\email{xuxuxuxuxuxjh@gmail.com}

\author{Changchang Yin}
\affiliation{%
  \institution{The Ohio State University Wexner Medical Center}
  \city{Columbus}
  \state{Ohio}
  \country{USA}
}
\email{Changchang.Yin@osumc.edu}

\author{Odysseas Chatzipanagiotou}
\affiliation{
\institution{The Ohio State University Wexner Medical Center}
  \city{Columbus}
  \state{Ohio}
  \country{USA}
}
\email{chatzipanagiotou.1@osu.edu}

\author{Diamantis Tsilimigras}
\affiliation{
\institution{The Ohio State University Wexner Medical Center}
  \city{Columbus}
  \state{Ohio}
  \country{USA}
}
\email{Diamantis.Tsilimigras@osumc.edu}

\author{Kevin Clear}
\affiliation{
\institution{The Ohio State University Wexner Medical Center}
  \city{Columbus}
  \state{Ohio}
  \country{USA}
}
\email{Kevin.Clear@osumc.edu}

\author{Bingsheng Yao}
\affiliation{
    \institution{Northeastern University}
    \city{Boston}
    \country{USA}}
\email{b.yao@northeastern.edu}

\author{Dakuo Wang}
\affiliation{
    \institution{Northeastern University}
    \city{Boston}
    \country{USA}}
\email{d.wang@northeastern.edu}

\author{Timothy Pawlik}
\affiliation{
\institution{The Ohio State University Wexner Medical Center}
  \city{Columbus}
  \state{Ohio}
  \country{USA}
}
\email{Tim.Pawlik@osumc.edu}

\author{Ping Zhang}
\affiliation{%
  \institution{The Ohio State University}
  \city{Columbus}
  \state{Ohio}
  \country{USA}
}
\email{zhang.10631@osu.edu}

\renewcommand{\shortauthors}{Xu et al.}

\begin{abstract}
Surgical site infection (SSI) is one of the most common and costly healthcare-associated infections and and surgical wound care remains a significant clinical challenge in preventing SSIs and improving patient outcomes. 
While recent studies have explored the use of deep learning for preliminary surgical wound screening, progress has been hindered by concerns over data privacy and the high costs associated with expert annotation. Currently, no publicly available dataset or benchmark encompasses various types of surgical wounds, resulting in the absence of an open-source Surgical-Wound screening tool.
To address this gap: (1) we present SurgWound, the first open-source dataset featuring a diverse array of surgical wound types. It contains 697 surgical wound images annotated by 3 professional surgeons with eight fine-grained clinical attributes. (2) Based on SurgWound, we introduce the first benchmark for surgical wound diagnosis, which includes visual question answering (VQA) and report generation tasks to comprehensively evaluate model performance. (3) Furthermore, we propose a three-stage learning framework, WoundQwen, for surgical wound diagnosis. In the first stage, we employ five independent MLLMs to accurately predict specific surgical wound characteristics. In the second stage, these predictions serve as additional knowledge inputs to two MLLMs responsible for diagnosing outcomes, which assess infection risk and guide subsequent interventions. In the third stage, we train a MLLM that integrates the diagnostic results from the previous two stages to produce a comprehensive report. This three-stage framework can analyze detailed surgical wound characteristics and provide subsequent instructions to patients based on surgical images, paving the way for personalized wound care, timely intervention, and improved patient outcomes.
\end{abstract}


\maketitle

\section{Introduction}


\begin{figure}[t!]
  \centering
  \includegraphics[width=\columnwidth]{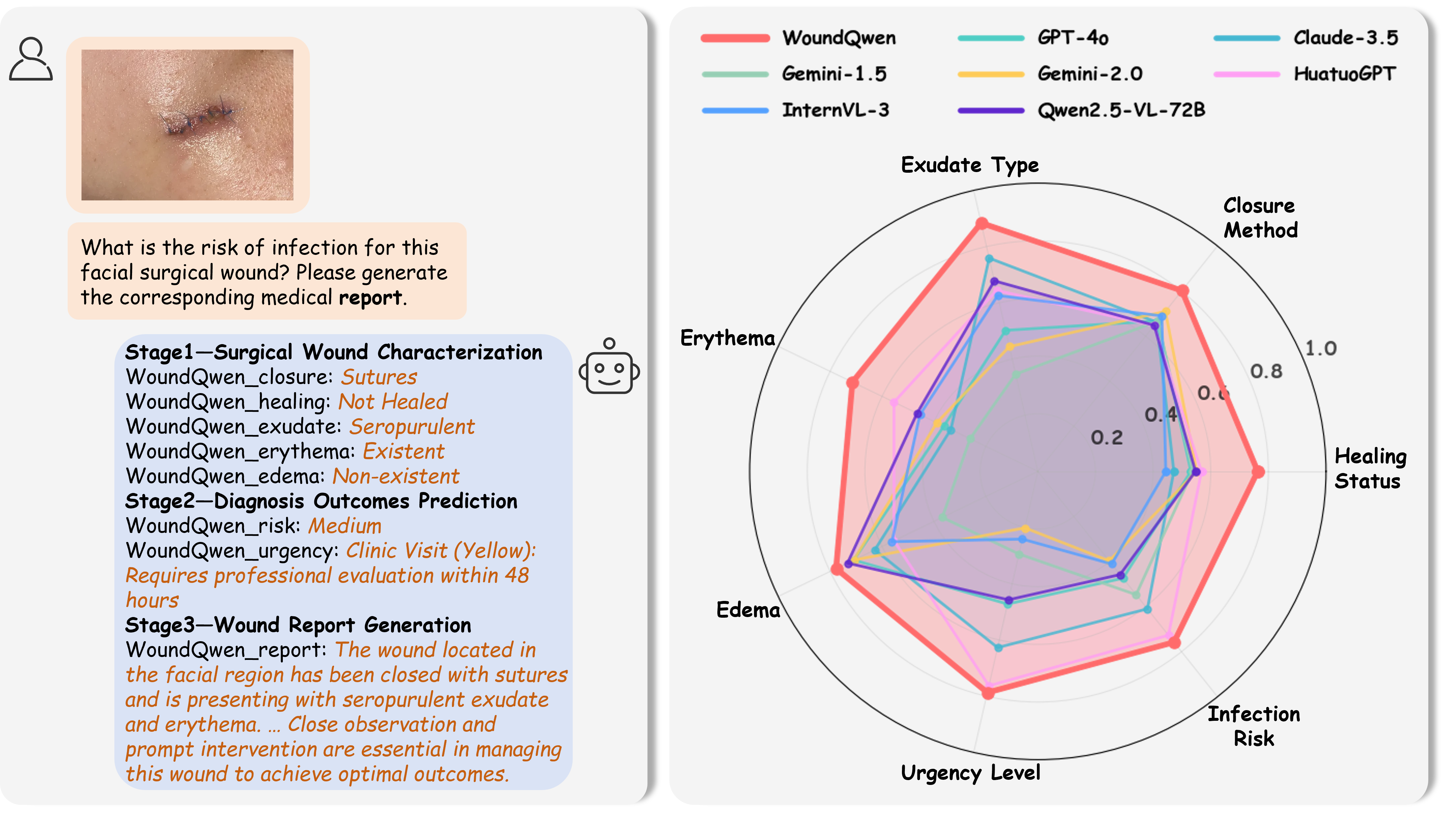}
  \caption{(1). WoundQwen is the first three-stage framework for surgical wound diagnosis.The left figure illustrates the WoundQwen workflow. (2). The right figure compares WoundQwen with existing MLLMs on the VQA task, using weighted‑F1 score as the metric.}
  \label{}
\end{figure}

Surgical site infection (SSI) is defined as an infection occurring within 30 days after a surgical procedure, affecting either the incision or the deep tissues at the operation site \cite{owens2008surgical}. SSI is among the most common and costly healthcare-associated infections \cite{iskandar2019highlighting}. It not only increases morbidity and mortality but also imposes a substantial economic burden and adversely affects patients’ quality of life \cite{badia2017impact}. 
Traditional surgical wound care is a resource-intensive process that often requires patients to make frequent in-person visits to healthcare facilities \cite{borst2025woundaissist}. This not only demands significant time and financial investment from patients but also places a considerable burden on healthcare systems. In contrast, the rapid advancements in artificial intelligence (AI) have enabled remote surgical wound monitoring to address these challenges. 

Although remote digital surgical wound monitoring is still in the early stages \cite{mclean2023readiness}, existing studies on emerging high-quality evidence \cite{mclean2021remote}, real-world implementations \cite{rochon2023image, rochon2024post}, and ongoing studies \cite{sandy2022non, rochon2024wound} have shown that AI can offer substantial benefits—including early detection of SSI, fewer clinic visits, and reduced costs and time burdens for patients \cite{rochon2025remote}. Motivated by the potential benefits of remote surgical wound monitoring, several recent studies, such as \cite{shenoy2018deepwound, chunlin2023deep, mclean2021remote, mclean2025multimodal}, have explored various AI models and approaches to enable preliminary diagnosis of surgical wounds. These efforts aim to reduce clinicians' workload and assist them in conducting remote assessments and follow-up care.

However, surgical wound diagnosis still faces several critical challenges. First, although a few open-source datasets exist for general wound analysis (e.g., diabetic foot ulcers \cite{cassidy2021dfuc, alzubaidi2020dfu_qutnet}), there is currently no publicly available dataset or benchmark specifically designed for diverse surgical wounds, as noted in the scoping review \cite{tabja2024machine}. Consequently, most existing studies \cite{mclean2021remote, mclean2023evaluation, mclean2025multimodal} rely on private datasets, which makes it difficult to fairly compare the performance of different methods and hinders reproducibility. We also investigate other publicly available AI tools, like wound-related apps \cite{AuxilliumHealth2025_WoundTele, Medline2025_SkinHealth, HealthyIO2024_Minuteful, RedScar2024_App} on Google Play and the App Store. However, Table 16 shows that none of these apps functioned properly.
Second, most existing approaches \cite{fletcher2021use, fletcher2021use2, wu2020unified, mclean2025multimodal} predominantly utilize CNN-based models that are limited to image-domain classification tasks. These models cannot incorporate patient-provided textual information such as wound location, making it difficult to support intelligent, multimodal, and patient-centric diagnostic processes (e.g., delivering timely and understandable instructions to patients).




In this paper, we present SurgWound, a high-quality surgical wound dataset constructed from a large collection of real-world wound images sourced from various social media platforms. The images were rigorously filtered by both AI and human experts to exclude non-surgical wounds and images of poor visual quality. Subsequently, each surgical wound image was annotated by three experienced surgeons with 8 fine-grained clinical attributes (e.g., wound location, healing status, infection risk). We provide a comprehensive analysis of the dataset and introduce SurgWound-Bench, a benchmark consisting of two tasks: visual question answering (VQA) and medical report generation.To address the inherent data imbalance in surgical wound diagnosis, we evaluate model performance on the VQA task using seven metrics: accuracy, precision, recall, F1 score, micro-F1, macro-F1, and weighted F1. To better assess the model’s report generation capability, we employ three metrics—BLEU, ROUGE, and BERTScore—as evaluation measures for the report generation task.

To facilitate interactive, multimodal surgical wound diagnosis, we introduce WoundQwen, the first three-stage diagnostic framework based on multimodal large language models (MLLMs). In Stage 1, we individually train five Qwen-based models using LoRA SFT to predict five surgical wound characteristics :\textit{Healing Status}, \textit{Closure Method}, \textit{Exudate Type}, \textit{Erythema}, and \textit{Edema}. In Stage 2, we incorporate the predicted surgical wound characteristics from Stage 1, along with the patient's known wound location, as additional contextual cues. \textit{WoundQwen\_risk} and \textit{WoundQwen\_urgency} leverage these cues to improve the accuracy of diagnosing \textit{Infection Risk} and \textit{Urgency Level}. In Stage 3, integrating the surgical wound prediction results from the first two stages, the trained \textit{WoundQwen\_report} enables high-quality report generation.

In summary, our contributions are as follows:
\begin{enumerate}
\item We introduce SurgWound, the first open-source dataset for surgical wound analysis across multiple procedure types. SurgWound comprises 697 surgical wound images, each annotated by surgical experts at The Ohio State University Wexner Medical Center (OSWUMC). Each image is accompanied by high-quality labels covering six surgical wound characteristic attributes and two diagnostic outcomes attributes.

\item We establish SurgWound-Bench, the first multimodal benchmark for surgical wound analysis, which includes two tasks: SurgWound-VQA and SurgWound-Report.

\item We designed a three-stage diagnostic framework, WoundQwen, the first MLLM-based framework for surgical wound diagnosis and personalized wound care instructions. WoundQwen achieves state-of-the-art performance on our SurgWound-Bench benchmark.

\item We release SurgWound, SurgWound-Bench, and WoundQwen as open-source resources to support future research and development in the surgical wound infection challenge.
\end{enumerate}

\noindent \textbf{Code and Docs}: https://github.com/xuxuxuxuxuxjh/SurgWound

\noindent \textbf{Dataset and Benchmark download}: 

\noindent https://huggingface.co/datasets/xuxuxuxuxu/SurgWound

\noindent SurgWound is also under submission to PhysioNet.

\noindent \textbf{License}: The dataset is released under the CC BY-SA 4.0 license.


\section{Related Work}

\subsection{Surgical Wound Infections}

Surgical wound infections pose a substantial challenge in clinical practice, highlighting the need for advanced technological solutions to enable continuous and effective monitoring. To address this issue, Deepwound \cite{shenoy2018deepwound} proposed an approach that combines mobile phone imaging with deep neural networks to classify surgical wounds, enabling patients to track wound healing and recovery from the comfort of their homes. Building on this vision, TWIST \cite{mclean2021remote} conducted the first comprehensive randomized controlled trial in this domain, demonstrating the feasibility, safety, and clinical efficacy of remote postoperative wound monitoring. Subsequently, INROADE \cite{mclean2023evaluation} extended this line of research with more comprehensive experiments and evaluations focused on gastrointestinal surgical wounds. With the success of ResNet-50 \cite{he2016deep} in image classification tasks, it has become a widely adopted backbone in surgical wound infection classification. Zhao et al.\cite{chunlin2023deep} evaluated four ResNet variants for surgical wound classification and identified SE-ResNet101 as the optimal model for remote monitoring, owing to its superior average accuracy. Fletcher et al \cite{fletcher2021use, fletcher2021use2}. utilized both thermal cameras and Android tablets to collect thermal and RGB images of cesarean section wounds, applying a ResNet-50-based model for infection classification. Wu et al. \cite{wu2020unified} proposed a unified framework for automatic wound infection detection, encompassing both an annotation tool and a ResNet-50-based classification model. 
To incorporate patient-specific clinical information, McLean et al. \cite{mclean2025multimodal} developed a multimodal machine learning model that extracts features from electronic health records (EHR) using an MLP, and from surgical wound images using VGG-19 \cite{simonyan2014very}. The text and image features are then fused to produce the final infection classification.

\subsection{Medical Multimodal Large Language Model}

With the emergence of powerful general MLLMs such as GPT-4 \cite{brown2020language, ouyang2022training} and ChatGPT \cite{achiam2023gpt}, which exhibit strong representation capabilities , a number of medical-domain MLLMs have also been proposed, including HuatuoGPT-Vision \cite{chen2024huatuogpt}, LLaVA-Med \cite{li2023llava}, Med-Flamingo \cite{moor2023med}, and MedBLIP \cite{chen2024medblip}.By leveraging the architecture of general MLLMs (e.g., Qwen2-VL \cite{wang2024qwen2}, LLaVA \cite{liu2023visual}, BLIP \cite{li2022blip}) and fine-tuning them on domain-specific image–text datasets, these models have demonstrated strong capabilities in medical visual understanding tasks. \quad In the specific filed of wound infection classification, SCARWID \cite{busaranuvong2025explainable} introduced a novel approach that leverages GPT-4o to generate captions for diabetic foot ulcer (DFU) images. These image–caption pairs are then used to train a BLIP-based model,which demonstrates strong performance in DFU infection classification.

\subsection{Wound Diagnosis APPs}

In recent years, there has been a surge in mobile applications designed to assist with wound monitoring and diagnosis. Appendix C.3, Table~\ref{tab:wound-apps}, summarizes 19 wound-related apps available on Google Play and the App Store. Most of these applications serve general wound care purposes, with only RedScar \cite{RedScar2024_App} targeting a specific surgical domain—abdominal wounds. Eleven of the apps require internal accounts, which can only be obtained by physicians affiliated with partner hospitals. The remaining eight apps, although accessible without internal credentials, fail to function properly due to issues such as frequent crashes, persistent error messages, connection failures, or prolonged waiting times. A key focus of modern wound apps is the integration of AI-driven tools. Applications like Minuteful for Wound \cite{HealthyIO2024_Minuteful}, Medline Skin Health \cite{Medline2025_SkinHealth}, and InteliWound \cite{Synergy2025_InteliWound} offer automated wound measurement functionalities, enhancing objectivity and efficiency in clinical workflows. However, only a small subset—such as WoundTele \cite{AuxilliumHealth2025_WoundTele}, PointClickCare \cite{PointClickCare2025_SkinAndWound}, and Medline Skin Health \cite{Medline2025_SkinHealth}—support AI-based infection risk or complication assessments.

\section{SurgWound Dataset and Benchmark}

In this section, we introduce the developed SurgWound Dataset and constructed benchmark.

\subsection{Overview}

SurgWound consists of a total of 686 wound images, which are divided into 480 images for training, 173 images for testing, and 69 images for validation. It is collected from multiple social media and online platforms, including \textit{RedNote}, \textit{Twitter}, \textit{Facebook}, \textit{Instagram}, and \textit{Reddit}. The dataset covers a wide range of wound characteristics and clinical attributes, providing a comprehensive resource for the development and evaluation of automated wound assessment models.

In SurgWound, each surgical wound image is annotated with 8 structured clinical labels, independently provided by three experienced professional surgeons to ensure labeling accuracy and clinical relevance. These 8 structured lables include 6 attributes capturing wound-specific features—namely \textit{Location}, \textit{Healing Status}, \textit{Closure Method}, \textit{Exudate Type}, \textit{Erythema Presence}, and \textit{Edema Presence}—as well as 2 clinical assessment labels: \textit{Infection Risk Assessment} and the \textit{Urgency Level} based treatment recommendation. The detailed summary of the annotations and their distribution in the SurgWound dataset is presented in \autoref{tab:dataset}. It is worth noting the following special cases in the annotation scheme: (1). The \textit{Others} option under \textit{Location} indicates that the wound location is identifiable but does not belong to common anatomical categories such as abdomen, ankle, or manus. (2) The \textit{Uncertain} label appears in several categories-\textit{Location}, \textit{Closure Method}, \textit{Exudate Type}, \textit{Erythema}, and \textit{Edema}—when the attribute could not be reliably determined based on the wound image alone.

\begin{table}
\caption{Annotation Categories and Data Distribution in the SurgWound Dataset}
\label{tab:dataset}
\centering
\begin{tabular}{p{2.3
cm} c p{2.3cm} c}
\hline
\textbf{Statistics} & \textbf{Number} & \textbf{Statistics} & \textbf{Number}  \\
\hline
 \textbf{Location} & 686 (100\%) & \textbf{Healing Status} & 686 (100\%) \\
-- Abdomen & 112 (16.3\%) & -- Healed & 406 (59.2\%) \\
-- Ankle & 72 (10.5\%) & -- Not Healed & 280 (40.8\%) \\
-- Facial region & 66 (9.6\%) & \textbf{Closure Method} & 686 (100\%) \\
-- Manus & 35 (5.1\%) & -- Invisible & 115 (16.8\%) \\
-- Patella & 35 (5.1\%) & -- Staples & 8 (1.2\%) \\
-- Cervical region & 26 (3.8\%) & -- Sutures & 350 (51.0\%) \\
-- Other & 178 (26\%) & -- Adhesives & 19 (2.8\%) \\
-- Uncertain & 162 (23.6\%) & -- Uncertain & 194 (28.3\%) \\
\hline
\textbf{Exudate Type} & 686 (100\%) & \textbf{Erythema} & 686 (100\%) \\
-- Non-existent & 531 (77.4\%) & -- Non-existent & 469 (68.4\%) \\
-- Purulent & 19 (2.8\%) & -- Existent & 191 (27.8\%) \\
-- Serous & 36 (5.3\%) & -- Uncertain & 26 (3.8\%) \\
-- Sanguineous & 29 (4.2\%) & \textbf{Edema} & 686 (100\%) \\
-- Seropurulent & 11 (1.6\%) & -- Non-existent & 473 (69.0\%) \\
-- Uncertain & 60 (8.7\%) & -- Existent & 81 (11.8\%) \\
& & -- Uncertain & 132 (19.2\%) \\
\hline
\textbf{Urgency Level} & 686 (100\%) & \textbf{Infection Risk} & 686 (100\%) \\
-- Home Care & 600 (87.5\%) & -- Low & 573 (83.5\%) \\
-- Clinic Visit & 68 (10.9\%) & -- Medium & 90 (13.1\%) \\
-- Emergency Care & 18 (2.6\%) & -- High & 23 (3.4\%) \\
\hline
\end{tabular}
\end{table}

\subsection{Dataset Construction}

\begin{figure*}[!t]  
  \centering
  \includegraphics[width=0.8\textwidth]{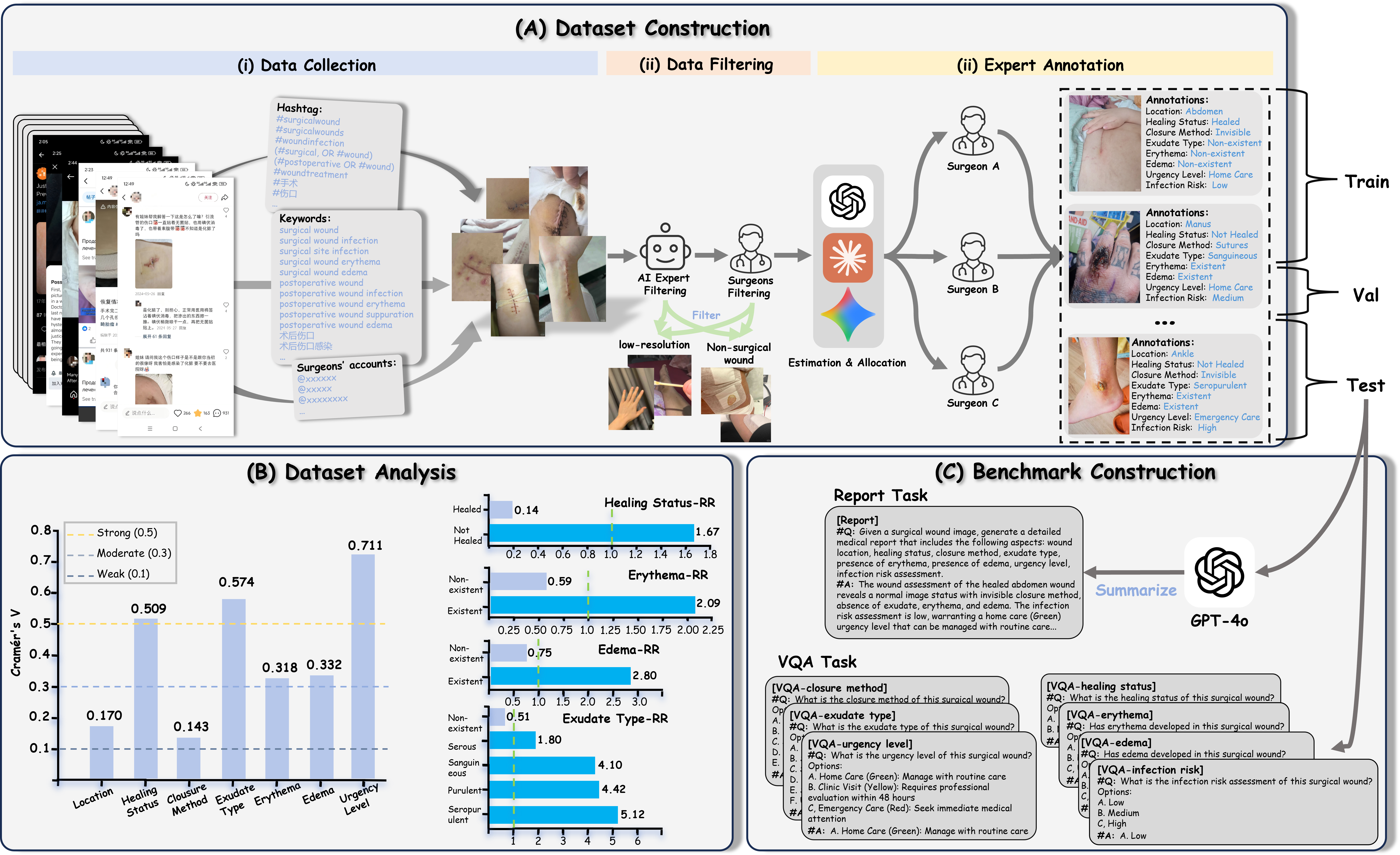}
  \caption{(A) Dataset Construction Pipeline: Includes three stages—data collection, data filtering, and expert annotation. (B)  Dataset Analysis: Examined associations between surgical wound characteristics and diagnostic outcomes via chi‑square tests; additionally used relative risk analysis to assess relationships with elevated infection risk. (C) Benchmark Construction Pipeline: Outlines creation of the report generation task and the VQA task.}
  \Description{}
  \label{fig:data_construction}
\end{figure*}

In this section, we introduce the SurgWound dataset construction pipeline, as shown in \autoref{fig:data_construction} (A).

\noindent \textbf{Data Collection.} Medical data present unique privacy challenges, particularly in the domain of surgical wound monitoring. To address potential ethical and privacy concerns while still acquiring large-scale, real-world data, we collect surgical wound images from publicly available content on social media platforms. Specifically, we utilize a collection of domain-specific hashtags (e.g., \textit{\#surgicalwoundinfection}) and keywords (e.g., \textit{postoperative wound}) to extract relevant content from various platforms, including \textit{RedNote}, \textit{Twitter}, \textit{Facebook}, \textit{Instagram}, and \textit{Reddit}. In addition, we further expand the dataset by collecting images specifically from the social media accounts of surgeons and other medical professionals, where postoperative wound cases are often shared for educational or awareness purposes.

\noindent \textbf{Data Filtering.} We perform a two-stage filtering process involving both AI expert and human expert review to ensure that only high-quality images containing visible surgical wounds are included in the dataset.
First, we leverage GPT-4o as an AI expert to automatically assess whether an image depicts a clear surgical wound, filtering out low-resolution images or those lacking any wound-related content.
Subsequently, three surgeons serve as human experts to manually review the remaining images and exclude any that low-resolution or do not depict authentic surgical wounds.

\noindent \textbf{Expert Annotation.} To ensure high-quality annotation while optimizing expert effort, we estimate the difficulty level of each image based on the predicted Risk Level outputs from three MLLMs: GPT-4o, Claude 3.5, and Gemini 2. Images for which all three models consistently predict a Low risk level are considered low-difficulty, while those with inconsistent predictions or predicted as Medium or High risk are categorized as high-difficulty. For low-difficulty cases, a single surgeon is randomly assigned to perform the annotation. For high-difficulty cases, the image is independently annotated by two randomly assigned surgeons. If any disagreement arises between their annotations, a third surgeon is introduced to review both sets of annotations and make the final decision.

\subsection{Dataset Analysis}

The SurgWound dataset includes six attributes describing the characteristics of surgical wounds, along with two attributes representing diagnostic outcomes.
\quad First, we analyzed the correlation between the two diagnostic outcomes: \textit{infection risk} and \textit{urgency level}. According to the statistical data, we found that for the vast majority of surgical wounds, infection risk corresponds directly with urgency level—that is, low risk corresponds to Home Care, medium risk corresponds to Clinic Visit, and high risk corresponds to Emergency Care. However, a small subset (26 cases) shows a slight deviation, where medium risk corresponds to Home Care. Through surgeons review, this discrepancy was attributed to surgical wound presentations exhibiting medium-risk features but displaying only mild or localized signs of infection—insufficient to meet the thresholds for clinical evaluation.
\quad Second, we examined the relationships between the six wound characteristics and the final diagnostic result. Given the strong correlation observed between urgency level and infection risk, we decided to focus our analysis solely on infection risk. From a practical perspective in remote surgical wound monitoring, both medium and high infection risks necessitate clinical intervention. Therefore, we combined these two categories in our analysis to more accurately reflect the need for medical attention.

\noindent \textbf{Independence Analysis Using Chi-Square Test.
}
We assessed the relationships between 7 surgical wound attributes—\textit{Location}, \textit{Healing Status}, \textit{Closure Method}, \textit{Exudate Type}, \textit{Erythema Presence}, \textit{Edema Presence}, \textit{Urgency Level}—and \textit{Infection Risk} using the chi-square test of independence. The steps of the chi-square test are as follows:(1) Assumed null hypothesis $H_0$: attribute is independent of infection risk. (2) Calculated the $\chi^2$ and p-value. The p-values for the seven attributes are summarized below:
\begin{center}
\begin{tabular}{p{1.5cm} c p{2.5cm} c}
\hline
\textbf{Attribute} & \textbf{p-value} & \textbf{Attribute} & \textbf{p-value}  \\ 
\hline
Location & 0.0197 & Closure Method & 0.0182 \\
Erythema & <0.0001 & Healing Status & <0.0001 \\
Edema & <0.0001 & Exudate Type & <0.0001 \\
& & Urgency Level & <0.0001  \\
\hline
\end{tabular}
\end{center}

\noindent Since all p-values are below 0.05, we reject the null hypothesis  $H_0$ and conclude that each of the seven attributes is associated with \textit{Infection Risk}. (3) Computed Cramér’s V to quantify the strength of these associations. The computed Cramér’s V values are presented in Fig 1 (B). First, we further validated the strong correlation between Urgency Level and Infection Risk. Second, we concluded that \textit{Healing Status} and \textit{Exudate Type} have a strong correlation with the diagnostic outcomes, while \textit{Erythema} and \textit{Edema} demonstrate a moderate correlation, and \textit{Location} and \textit{Closure Method} exhibit a weaker correlation.

\noindent\textbf{Relative Risk Analysis of Surgical Wound Characteristics.}
Based on chi-square test results above, we found that certain attributes—\textit{Healing Status}, \textit{Exudate Type}, \textit{Erythema}, and \textit{Edema}—are significantly associated with Infection Risk. To further quantify which surgical wound characteristics contribute most to elevated risk (medium or high risk), we computed the Relative Risk (RR). Relative Risk is the ratio of the incidence of outcome (medium or high risk) in the exposed group (e.g., exhibiting a given surgical wound characteristic) versus the non-exposed group. Based on the Relative Risk results shown in \autoref{fig:data_construction} (B), we identified the following surgical wound characteristics as elevated risk factors: \textit{Not Healed}, \textit{Erythema}, \textit{Edema}, and exudate types including \textit{Serous}, \textit{Sanguineous}, \textit{Purulent}, and \textit{Seropurulent}, which were associated with significantly increased risk of infection.

\subsection{Benchmark Construction}

The SurgWound dataset is divided into training, validation, and test sets in a 7:1:2 ratio. The test set comprises 173 surgical wound images. Using these images and their corresponding annotations, we design two tasks: visual question answering (VQA) and report generation. The VQA task evaluates the model’s ability to accurately extract clinical attributes from surgical wound images in a question-and-answer format. The report generation task requires the model to produce a coherent and informative textual summary of the surgical wound condition, simulating the role of an intelligent assistant in clinical documentation and patient communication.

\noindent \textbf{Visual Question Answering (VQA) Task.} To create the multiple-choice QA pairs, we leverage the eight manually annotated attributes for each image. Each individual attribute generates a question-answer pair. Importantly, any annotated attribute marked as “Uncertain” is excluded to maintain the quality of the questions and answers. This ensures that only clear, unambiguous attribute values are used when forming the QA pairs.

\noindent \textbf{Report Generation Task.}
Previous studies generate medical reports by directly inputting images and prompts into advanced MLLMs, such as GPT-4o. However, general MLLMs tend to produce hallucinations or inaccurate content when applied to surgical wound reporting. \quad To address this issue, we adopt a two-step generation pipeline. First, we use GPT-4o to generate a structured medical report based on the surgical wound image and its eight annotated attributes. Then, each generated report is manually reviewed and revised by professional surgeons to ensure clinical accuracy and consistency.

\subsection{Evaluation Metrics}
\noindent \textbf{Visual Question Answering (VQA) Task.}
Due to the typical medical imaging imbalance issue with surgical wound data, we evaluate our models using a comprehensive set of metrics. In addition to Accuracy, we compute Precision, Recall, F1-score, Micro-F1, Macro-F1, and Weighted-F1 to better assess model performance under imbalanced conditions. 





\[
\text{Micro‑}F_1 = \frac{2\,\sum_{k} \mathrm{TP}_k}{2\,\sum_{k} \mathrm{TP}_k \;+\; \sum_{k} \mathrm{FP}_k \;+\; \sum_{k} \mathrm{FN}_k}
\]

\[
\text{Macro‑}F_1 = \frac{1}{K} \sum_{k=1}^{K} F_{1,k}
\]

\[
\text{Weighted‑}F_1 = \frac{1}{\sum_{k} n_k} \sum_{k=1}^{K} n_k \cdot F_{1,k}
\]

\noindent where \(K\) is the total number of classes, \(\mathrm{TP}_k\), \(\mathrm{FP}_k\), and \(\mathrm{FN}_k\) represent the true positives, false positives, and false negatives for class \(k\), respectively. \(F_{1,k}\) denotes the F1 score computed for class \(k\). Finally, \(n_k\) is the number of true instances (support) for class \(k\).

\noindent \textbf{Report Generation Task.}
To evaluate report generation performance, we utilize three metrics: BLEU \cite{papineni2002bleu}, ROUGE \cite{lin2004rouge}, and BERTScore \cite{zhang2019bertscore}. Specifically, BLEU includes BLEU-1 and BLEU-2, while ROUGE comprises ROUGE-1 and ROUGE-L.

\begin{figure}[!t]  
  \centering
  \includegraphics[width=\columnwidth]{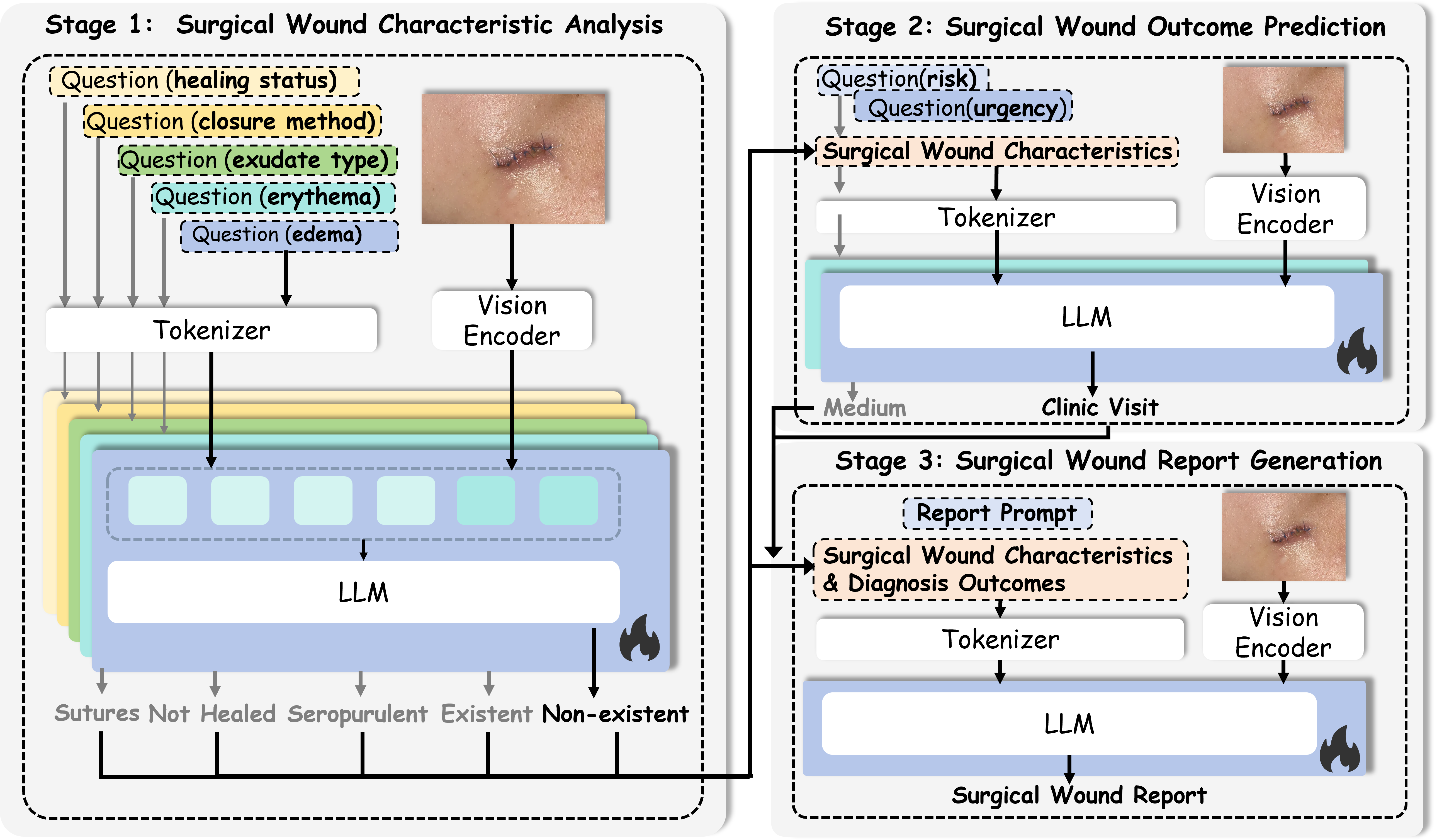}
  \caption{Three-Stage Diagnostic Framework.}
  \Description{}
  \label{fig:model}
\end{figure}

\section{Three-Stage Diagnostic Framework}
In the task of surgical wound diagnosis, the most critical outcomes are the \textit{Infection Risk} and the corresponding clinical action (\textit{Urgency Level}), as defined in our SurgWound-Bench benchmark. However, the other six surgical wound characteristics—Healing Status, Exudate Type, Erythema, Edema, Closure Method, and Wound Location—are equally important, as they serve as clinical evidence guiding the final diagnosis. These characteristics assist clinicians or AI models in making accurate decisions. To leverage these insights, we propose a three-stage diagnostic framework, as illustrated in \autoref{fig:model}. In the first stage, five of the six wound characteristics are predicted by specialized models: \textit{Healing Status}, \textit{Closure Method}, \textit{Exudate Type}, \textit{Erythema}, and \textit{Edema}. \textit{Location} is considered known clinical information and is not predicted. In the second stage, the wound image, together with the predicted characteristics and wound location, is input into two specialized models—WoundQwen\_risk for infection risk prediction and WoundQwen\_urgency for urgency level prediction. In the third stage, WoundQwen\_report utilizes the predictions from the first two stages along with the known location information to analyze images and generate a surgical wound report.

\subsection{Stage 1: Surgical Wound Characteristic Analysis}
In Stage 1, we utilize five separately trained MLLMs to predict five surgical wound characteristics—Healing Status, Closure Method, Exudate Type, Erythema, and Edema—from surgical wound images. These models employ the Qwen2.5-VL-7B \cite{bai2025qwen2} architecture, initialized with the pre-trained weights of HuatuoGPT-Vision-7B \cite{chen2024huatuogpt}, and are fine-tuned using supervised fine-tuning (SFT) with LoRA \cite{hu2022lora}. We freeze the entire vision encoder and only adapt the large language model layers. Specifically, LoRA adapters are applied to the attention modules (q\_proj, k\_proj, v\_proj, o\_proj) and the MLP modules (gate\_proj, up\_proj, down\_proj), while all other parameters remain frozen. We convert annotated images into VQA-style instruction pairs, each consisting of an image and a text question such as “\textit{What is the healing status of this surgical wound?}”, followed by the correct textual answer. The models are trained separately on this VQA instruction data to obtain five characteristic predictors, denoted as $\{\hat{y}_i\}_{i=1}^{5}$.

\subsection{Stage 2: Surgical Wound Outcome Prediction}
Building on the Stage 1 predictions, Stage 2 aggregates the outputs of the five characteristic models along with the known wound location. These are concatenated as additional knowledge inputs to a final Qwen-based model, referred to as \textit{WoundQwen\_risk} and \textit{WoundQwen\_urgency}, which predicts two diagnostic outcomes: \textit{Urgency Level} and \textit{Infection Risk}.

\[ 
\begin{alignedat}{2}
\hat{y}_\text{risk} &= \mathrm{WoundQwen\_risk}\bigl(\text{Image},\; \textit{Location},\; \{\hat{y}_i\}_{i=1}^{5}\bigr), \\[4pt]
\hat{y}_\text{urgency} &= \mathrm{WoundQwen\_urgency}\bigl(\text{Image},\; \textit{Location},\; \{\hat{y}_i\}_{i=1}^{5}\bigr).
\end{alignedat}
\]

\noindent where \(\{\hat{y}_i\}\) denotes the set of predicted surgical wound characteristics from Stage 1.

This cascade modeling approach, similar in concept to multistage multimodal frameworks in clinical prognosis, enables the final classifier to leverage detailed intermediate feature predictions to improve diagnostic accuracy. Furthermore, such a modular architecture enhances interpretability and allows for targeted optimization of each task independently. To validate the relevance of each surgical wound characteristic, we conducted chi-square tests and relative risk (RR) analyses. For more detailed information, please refer to Section 3.

In Stage 2, our MLLMs continue to be trained using LoRA-based supervised fine-tuning (SFT). For the training input text, we use initial questions supplemented with additional context describing surgical wound characteristics to meet the requirements of our second-stage model.

\subsection{Stage 3: Surgical Wound Report Generation}

In Stage 3, we incorporate the wound characteristics and diagnostic outcomes predicted in Stages 1 and 2 as structured contextual inputs to WoundQwen\_report, enriching its understanding beyond raw images. These contextual cues—such as Healing Status, Closure Method, Infection Risk, and Urgency Level—are concatenated with the image during prompt construction, guiding the model toward clinically relevant content synthesis. Through training on high-quality clinical image–report pairs, WoundQwen\_report learns to generate rich, coherent diagnostic reports that align with real-world standards. Evaluation demonstrates that it significantly outperforms baseline and alternative models on metrics such as BLEU, ROUGE, and BERTScore, indicating superior report quality, clinical coherence, and alignment with specialist-authored narratives.

\section{Experiments}

\subsection{Implementation Details}
All experiments were conducted on A100 GPUs using PyTorch. During supervised fine-tuning (SFT), we froze the vision encoder and applied LoRA only to the LLM layers. We set the per-device batch size to 1, accumulating gradients over 8 steps for an effective batch size of 8. The learning rate was \(1\times10^{-4}\) with a cosine scheduler and 10\% warmup. At each stage, the model achieving the best performance on the validation set is selected as the final model.

\noindent \textbf{MLLMs.}
We evaluated our proposed three-stage diagnostic framework on the SurgWound-Bench benchmark, comparing its performance against current state-of-the-art MLLMs, including both proprietary API models and leading open-source models:



\begin{enumerate}
    \item \textbf{API models:} GPT-4o, Claude-3.5-Sonnet, Gemini-1.5-Pro, Gemini-2.0-Flash
    \item \textbf{Open-source model:} HuatuoGPT-Vision-34B, Qwen2.5-VL-72B, InternVL3-78B-78B
\end{enumerate}

\begin{table*}[t]
  \centering
  \caption{Accuracy and Weighted-F1 Score Results in VQA Task. The upper section of the table presents the accuracy and weighted-F1 scores of current state-of-the-art MLLMs, while the lower section displays the accuracy and weighted-F1 scores of WoundQwen and its ablation studies. Note: F1 in the table refers to the weighted-F1 score.}
  \setlength{\tabcolsep}{2pt}

  \begin{tabular}{l*{8}{cc}}
  \toprule
    Model
      & \multicolumn{2}{c}{Closure Method}
      & \multicolumn{2}{c}{Healing Status}
      & \multicolumn{2}{c}{Exudate Type}
      & \multicolumn{2}{c}{Erythema}
      & \multicolumn{2}{c}{Edema}
      & \multicolumn{2}{c}{Urgency Level}
      & \multicolumn{2}{c}{Infection Risk} \\
    \cmidrule(lr){2-3}\cmidrule(lr){4-5}\cmidrule(lr){6-7}
    \cmidrule(lr){8-9}\cmidrule(lr){10-11}\cmidrule(lr){12-13}\cmidrule(lr){14-15}\cmidrule(lr){16-17}
          & ACC & F1 & ACC & F1 & ACC & F1 & ACC & F1 & ACC & F1 & ACC & F1 & ACC & F1 \\
    \midrule
    GPT‑4o   & 63.04  &0.668    & 60.00  &0.531     & 39.20   &0.502        & 40.46  &0.362   & 62.61   &0.709  & 39.42   &0.472    & 40.88      &0.474        \\
    Claude‑3.5    & 59.78  &0.662  & 57.14  &0.472    & 68.80    &0.759    & 39.69  &0.335   & 57.39 &0.628  & 57.74  &0.627   & 54.01  &0.610           \\
    Gemini‑1.5            & 60.87  &0.667   & 61.31 &0.537    & 23.20    &0.346        & 34.35  &0.261   & 26.96   &0.368  & 26.28   &0.294    & 46.72      &0.547       \\
    Gemini‑2.0      & 68.48  &0.710  & 60.58  &0.551    & 30.00    &0.444      & 43.51   &0.389  & 64.35   &0.709  & 12.41    &0.201   & 35.77  &0.397             \\
    HuatuoGPT-34B   & 61.96  &0.647   & 58.02   &0.568    & 52.00  &0.644   & 60.31  &0.554  & 49.96  &0.554   & 74.45   &0.763     & 67.15  &0.727         \\
    InternVL3-78B   & 63.04   &0.690  & 56.20   &0.444    & 51.02       &0.626   & 46.56  &0.454   & 46.96  &0.562   & 19.71   &0.241  & 35.77   &0.411  \\
    Qwen2.5‑VL‑72B        & 61.96   &0.648  & 60.15   &0.547       & 57.60   &0.678     & 47.33   &0.465  & 73.04  &0.733   & 37.96  &0.457    & 38.69 &0.458     \\
    
\toprule

    HuatuoGPT‑7B &75.00 & 0.671 &62.04 & 0.617 &16.00 & 0.235 &67.18 & 0.680 &56.52 & 0.616 &34.31 & 0.456 &61.31 & 0.661 \\
\makecell[l]{+ LoRA} &78.26 & 0.765 &67.15 & 0.674 &83.20 & 0.821 &\textbf{74.05} & \textbf{0.729} &75.65 & 0.722 &75.18 & 0.779 &74.45 & 0.749 \\
\textbf{WoundQwen} &\textbf{81.52} & \textbf{0.802} &\textbf{76.64} & \textbf{0.763} &\textbf{88.80} & \textbf{0.884} &70.99 & 0.714 &\textbf{76.52}& \textbf{0.777} &\textbf{82.35} & \textbf{0.822} &\textbf{83.21} & \textbf{0.845} \\
    \bottomrule
  \end{tabular}
  \label{tab:acc_summary}
\end{table*}

\subsection{VQA Task}
In the VQA task, we designed seven sub-tasks: recognition of five surgical wound characteristics (Healing Status, Exudate Type, Erythema, Edema, Closure Method) and two diagnostic outcomes (Infection Risk and Urgency Level). To comprehensively evaluate the model's capabilities—especially its performance on rare classes—we adopted seven evaluation metrics: Accuracy, Precision, Recall, F1-Score, Micro-F1, Macro-F1, and Weighted-F1. Table 2 presents the accuracy and weighted-F1 score results. Infection risk and urgency level results are shown in Tables 3 and 4, respectively, with additional results available in Appendix C.1. WoundQwen consistently demonstrates superior performance across all seven VQA sub-tasks, achieving substantially higher accuracy and Weighted-F1 scores than competing models. In particular, it excels in Exudate Type (ACC 88.80 \%, F1 0.884), Infection Risk (ACC 83.21 \%, Weighted-F1 0.845), and Urgency Level (ACC 82.35 \%, Weighted-F1 0.822). Its marked advantage in recognizing rare or low-frequency classes—such as Exudate Type, where the weighted-F1 score is 0.884 compared to below 0.68 for others—indicates strong robustness against class imbalance. GPT-4o and Qwen2.5-VL-72B perform well on the Edema sub-task but show average results in assessing surgical wound infection risk. Claude-3.5 performs acceptably on Exudate Type but underperforms on most other sub-tasks. HuatuoGPT-34B achieves high scores for Urgency Level and Infection Risk but remains inferior in recognizing surgical wound characteristics. The fact that WoundQwen leads in both accuracy and F1 score highlights not only its high overall correctness but also its balanced precision and recall, even for rare categories. As shown in Table 3, WoundQwen demonstrates stronger performance in medium- and high-risk diagnoses compared to other models. This suggests that its powerful multimodal reasoning capabilities and domain-specific fine-tuning are key contributors to its superior performance. Therefore, WoundQwen emerges as the best choice for surgical wound diagnosis applications.

\begin{table*}[t]  
  \centering
  \caption{Infection Risk VQA task performance}
  \begin{tabular}{lcccccc}
    \toprule
    Model & Low (P/R/F1) & Medium (P/R/F1) & High (P/R/F1) & Micro‑F1 & Macro‑F1 & Weighted‑F1 \\
    \midrule
    GPT‑4o & 0.933/0.368/0.528 & 0.159/0.778/0.264 & 0.000/0.000/0.000 & 0.412 & 0.264 & 0.474 \\
    Claude‑3.5 & 0.901/0.561/0.692 & 0.169/0.556/0.260 & 0.000/0.000/0.000 & 0.548 & 0.317 & 0.610 \\
    Gemini‑1.5 & 0.930/0.465/0.620 & 0.147/0.611/0.237 & 0.000/0.000/0.000 & 0.467 & 0.285 & 0.547 \\
    Gemini‑2.0 & 1.000/0.263/0.417 & 0.170/1.000/0.290 & 1.000/0.200/0.333 & 0.358 & 0.347 & 0.397 \\
    HuatuoGPT-34B & 0.930/0.702/0.800 & 0.270/0.556/0.364 & 0.333/0.400/0.364 & 0.692 & 0.509 & 0.727 \\
    InternVL3-78B & 0.943/0.289/0.443 & 0.156/0.778/0.259 & 0.167/0.400/0.235 & 0.358 & 0.312 & 0.411 \\
    Qwen2.5‑VL‑72B & 0.909/0.351/0.506 & 0.141/0.667/0.233 & 0.143/0.200/0.167 & 0.388 & 0.302 & 0.458 \\
    \textbf{WoundQwen} & 0.952/0.877/0.913 & 0.429/0.667/0.522 & 0.500/0.400/0.444 & \textbf{0.832} & \textbf{0.626} & \textbf{0.845} \\
    \bottomrule
  \end{tabular}
  \label{tab:infection_risk_assessment}
\end{table*}

\begin{table*}[t]  
  \centering
  \caption{Urgency Level VQA task performance}
  \begin{tabular}{lcccccc}
    \toprule
    Model & Home Care (P/R/F1) & Clinic Visit (P/R/F1) & Emergency Care (P/R/F1) & Micro‑F1 & Macro‑F1 & Weighted‑F1 \\
    \midrule
    GPT‑4o                      & 0.933/0.353/0.512 & 0.135/0.750/0.229 & 0.000/0.000/0.000 & 0.397          & 0.247          & 0.472           \\
    Claude‑3.5                  & 0.941/0.538/0.684 & 0.172/0.688/0.275 & 0.000/0.000/0.000 & 0.556          & 0.320          & 0.627           \\
    Gemini‑1.5                  & 0.957/0.185/0.310 & 0.124/0.875/0.217 & 0.000/0.000/0.000 & 0.263          & 0.176          & 0.294           \\
    Gemini‑2.0                  & 1.000/0.118/0.211 & 0.054/0.125/0.075 & 1.000/0.500/0.667 & 0.180          & 0.318          & 0.201           \\
    HuatuoGPT-34B                      & 0.892/0.832/0.861 & 0.083/0.062/0.071 & 0.333/1.000/0.500 & 0.767          & \textbf{0.477}          & 0.763           \\
    InternVL3-78B                   & 1.000/0.143/0.250 & 0.107/0.562/0.180 & 0.111/0.500/0.182 & 0.219          & 0.204          & 0.241           \\
    Qwen2.5‑VL‑72B             & 0.930/0.336/0.494 & 0.126/0.687/0.214 & 0.143/0.500/0.222 & 0.380          & 0.310          & 0.457           \\
    \textbf{WoundQwen}       & 0.907/0.907/0.907 & 0.294/0.312/0.303 & 0.000/0.000/0.000 & \textbf{0.832}          & 0.403          & \textbf{0.822 }          \\
    \bottomrule
  \end{tabular}
  \label{tab:urgency}
\end{table*}

\subsection{Report Generation Task}
We report BLEU, ROUGE, and BERTScore metrics to evaluate the performance of surgical wound report generation, as shown in Tab. 10. The results clearly demonstrate that the WoundQwen framework achieved the highest scores across all evaluation metrics. Specifically, WoundQwen’s BLEU-1 score is 0.457, significantly surpassing the second-place HuatuoGPT-34B’s score of 0.252. Its BLEU-2 score is 0.300, also markedly higher than those of other models. These results indicate that WoundQwen more accurately reproduces n-grams from the reference text when generating reports, thereby achieving superior BLEU scores. Regarding ROUGE metrics, WoundQwen also shows a clear advantage, with ROUGE-1 and ROUGE-L scores of 0.612 and 0.418, respectively. In comparison, other models generally achieve ROUGE-1 scores between 0.39 and 0.44, while their ROUGE-L scores are significantly lower than those of WoundQwen. This suggests that WoundQwen’s generated reports have a higher content overlap with the reference reports. Under the BERTScore metric, WoundQwen attains a score of 0.917, again outperforming other models, with HuatuoGPT-34B in second place at 0.870. This indicates that WoundQwen’s outputs are semantically closer to the reference text. By contrast, general-purpose large models such as GPT-4o, Claude-3.5, and Gemini-1.5/2.0 generally score lower, indicating that their performance on this task is inferior to that of task-specific trained models.

\begin{table}[h]
\centering
\caption{Ablation Experiment}
  \begin{tabular}{l*{2}{cc}}
  \toprule
    Model
      & \multicolumn{2}{c}{Urgency Level}
      & \multicolumn{2}{c}{Infection Risk} \\
    \cmidrule(lr){2-3}\cmidrule(lr){4-5}
          & ACC & F1 & ACC & F1  \\
    \midrule
WoundQwen  &82.35 & 0.822 &83.21 & 0.845 \\
\makecell[l]{+ GT Cues} & \textbf{88.32} & \textbf{0.888} &\textbf{85.40} & \textbf{0.861} \\
\toprule
GPT‑4o &39.42 & 0.247 &40.88 & 0.474 \\
\makecell[l]{+ GT Cues}  &\textbf{70.80} & \textbf{0.759} &\textbf{72.99} & \textbf{0.770} \\
Claude‑3.5  &57.74 & 0.627 &54.01 & 0.610 \\
\makecell[l]{+ GT Cues}  &\textbf{61.31} & \textbf{0.681} &\textbf{76.64} & \textbf{0.798} \\
\bottomrule
\end{tabular}
\end{table}

\begin{table}[h]
  \centering
  \caption{Result in Report Generation Task}
  \begin{tabular}{p{2.2cm} p{0.7cm} p{0.7cm} p{1.0cm} p{1.0cm} p{0.9cm}}
    \toprule
    Model & BLEU‑1 & BLEU‑2  & ROUGE‑1 & ROUGE‑L & \makecell{BERT\\Score} \\
    \midrule
    GPT‑4o          & 0.160  & 0.072  & 0.412     & 0.213   & 0.852 \\
    Claude‑3.5      & 0.170  & 0.073  & 0.391    & 0.205   & 0.854 \\
    Gemini‑1.5      & 0.169  & 0.069  & 0.397    & 0.199   & 0.844 \\
    Gemini‑2.0      & 0.214  & 0.090   & 0.428    & 0.226   & 0.846 \\
    HuatuoGPT-34B          & 0.252 & 0.108  & 0.409  & 0.195 & 0.870 \\
    InternVL3-78B       & 0.161 & 0.072  & 0.351   & 0.180   & 0.838 \\
    Qwen2.5‑VL‑72B  & 0.234  & 0.112   & 0.435    & 0.231   & 0.874 \\
    \textbf{WoundQwen} & \textbf{0.457} & \textbf{0.300} & \textbf{0.612} & \textbf{0.418} & \textbf{0.917} \\
    \bottomrule
  \end{tabular}
\end{table}
\subsection{Ablation Experiment}

To validate the effectiveness of our three-stage diagnostic framework, we conducted three experiments: (1) evaluating the HuatuoGPT-7B base model and the single-stage LoRA-SFT training results; (2) comparing the diagnostic outcomes of WoundQwen\_risk and WoundQwen\_urgency using either the ground-truth surgical wound characteristics or the stage 1 predicted characteristics from WoundQwen as context; and (3) assessing the diagnostic performance of GPT-4o and Claude-3.5 when provided with ground-truth surgical wound characteristics as additional context. The results of Experiment (1) are presented in Table 2, while the results of Experiments (2) and (3) are shown in Table 5. After LoRA-SFT fine-tuning, HuatuoGPT-7B showed some improvement but remained significantly inferior to our three-stage diagnostic framework. GPT-4o and Claude-3.5 also demonstrated gains when ground-truth cues were added as context. Finally, because a gap remains between the stage-1 predicted features and the ground-truth features, WoundQwen\_risk and WoundQwen\_urgency performed better when using ground-truth features as context. Overall, our proposed three-stage diagnostic framework demonstrates strong performance on the surgical wound diagnosis task.




\section{Conclusion}
In this paper, we introduce SurgWound, the first open-source dataset for surgical wound analysis across multiple procedure types. By collecting surgical wound images from social media platforms, applying AI and human expert filtering, and obtaining annotations from three professional surgeons, we have constructed a dataset annotated with eight fine-grained clinical attributes, addressing a critical gap in the field. To better evaluate model performance in surgical wound diagnosis, we propose a benchmark—SurgWound-Bench—which includes both a VQA task and a report generation task. SurgWound-Bench is the first benchmark for surgical wound diagnosis and is poised to establish a foundation for future advancements in this domain. Finally, we design a three-stage diagnostic framework comprising surgical wound characteristic analysis, outcome prediction, and report generation. Our framework outperforms current state-of-the-art multimodal large language models on seven VQA sub-tasks and the report generation task. Extensive experiments provide key insights into both the dataset and the model’s diagnostic capabilities. We believe SurgWound-Bench represents a significant advancement in the field of AI-based surgical wound diagnosis.

\bibliographystyle{ACM-Reference-Format}
\bibliography{sample-base}


\begin{thebibliography}{45}


\ifx \showCODEN    \undefined \def \showCODEN     #1{\unskip}     \fi
\ifx \showISBNx    \undefined \def \showISBNx     #1{\unskip}     \fi
\ifx \showISBNxiii \undefined \def \showISBNxiii  #1{\unskip}     \fi
\ifx \showISSN     \undefined \def \showISSN      #1{\unskip}     \fi
\ifx \showLCCN     \undefined \def \showLCCN      #1{\unskip}     \fi
\ifx \shownote     \undefined \def \shownote      #1{#1}          \fi
\ifx \showarticletitle \undefined \def \showarticletitle #1{#1}   \fi
\ifx \showURL      \undefined \def \showURL       {\relax}        \fi
\providecommand\bibfield[2]{#2}
\providecommand\bibinfo[2]{#2}
\providecommand\natexlab[1]{#1}
\providecommand\showeprint[2][]{arXiv:#2}

\bibitem[Achiam et~al\mbox{.}(2023)]%
        {achiam2023gpt}
\bibfield{author}{\bibinfo{person}{Josh Achiam}, \bibinfo{person}{Steven Adler}, \bibinfo{person}{Sandhini Agarwal}, \bibinfo{person}{Lama Ahmad}, \bibinfo{person}{Ilge Akkaya}, \bibinfo{person}{Florencia~Leoni Aleman}, \bibinfo{person}{Diogo Almeida}, \bibinfo{person}{Janko Altenschmidt}, \bibinfo{person}{Sam Altman}, \bibinfo{person}{Shyamal Anadkat}, {et~al\mbox{.}}} \bibinfo{year}{2023}\natexlab{}.
\newblock \showarticletitle{Gpt-4 technical report}.
\newblock \bibinfo{journal}{\emph{arXiv preprint arXiv:2303.08774}} (\bibinfo{year}{2023}).
\newblock


\bibitem[Alzubaidi et~al\mbox{.}(2020)]%
        {alzubaidi2020dfu_qutnet}
\bibfield{author}{\bibinfo{person}{Laith Alzubaidi}, \bibinfo{person}{Mohammed~A Fadhel}, \bibinfo{person}{Sameer~R Oleiwi}, \bibinfo{person}{Omran Al-Shamma}, {and} \bibinfo{person}{Jinglan Zhang}.} \bibinfo{year}{2020}\natexlab{}.
\newblock \showarticletitle{DFU\_QUTNet: diabetic foot ulcer classification using novel deep convolutional neural network}.
\newblock \bibinfo{journal}{\emph{Multimedia Tools and Applications}} \bibinfo{volume}{79}, \bibinfo{number}{21} (\bibinfo{year}{2020}), \bibinfo{pages}{15655--15677}.
\newblock


\bibitem[{Auxillium Health LLC}(2025)]%
        {AuxilliumHealth2025_WoundTele}
\bibfield{author}{\bibinfo{person}{{Auxillium Health LLC}}.} \bibinfo{year}{2025}\natexlab{}.
\newblock \bibinfo{title}{WoundTele: AI Wound Care (Android App)}.
\newblock \bibinfo{howpublished}{\url{https://play.google.com/store/apps/details?id=com.auxilliumhealth.woundTele}}.
\newblock
\newblock
\shownote{Accessed: 2025‑07‑21; AI‑powered wound image analysis, risk scoring, remote consultation, and clinician dashboard integration}.


\bibitem[Badia et~al\mbox{.}(2017)]%
        {badia2017impact}
\bibfield{author}{\bibinfo{person}{JM Badia}, \bibinfo{person}{AL Casey}, \bibinfo{person}{N Petrosillo}, \bibinfo{person}{PM Hudson}, \bibinfo{person}{SA Mitchell}, {and} \bibinfo{person}{C Crosby}.} \bibinfo{year}{2017}\natexlab{}.
\newblock \showarticletitle{Impact of surgical site infection on healthcare costs and patient outcomes: a systematic review in six European countries}.
\newblock \bibinfo{journal}{\emph{Journal of Hospital Infection}} \bibinfo{volume}{96}, \bibinfo{number}{1} (\bibinfo{year}{2017}), \bibinfo{pages}{1--15}.
\newblock


\bibitem[Bai et~al\mbox{.}(2025)]%
        {bai2025qwen2}
\bibfield{author}{\bibinfo{person}{Shuai Bai}, \bibinfo{person}{Keqin Chen}, \bibinfo{person}{Xuejing Liu}, \bibinfo{person}{Jialin Wang}, \bibinfo{person}{Wenbin Ge}, \bibinfo{person}{Sibo Song}, \bibinfo{person}{Kai Dang}, \bibinfo{person}{Peng Wang}, \bibinfo{person}{Shijie Wang}, \bibinfo{person}{Jun Tang}, {et~al\mbox{.}}} \bibinfo{year}{2025}\natexlab{}.
\newblock \showarticletitle{Qwen2. 5-vl technical report}.
\newblock \bibinfo{journal}{\emph{arXiv preprint arXiv:2502.13923}} (\bibinfo{year}{2025}).
\newblock


\bibitem[Borst et~al\mbox{.}(2025)]%
        {borst2025woundaissist}
\bibfield{author}{\bibinfo{person}{Vanessa Borst}, \bibinfo{person}{Anna Riedmann}, \bibinfo{person}{Tassilo Dege}, \bibinfo{person}{Konstantin M{\"u}ller}, \bibinfo{person}{Astrid Schmieder}, \bibinfo{person}{Birgit Lugrin}, {and} \bibinfo{person}{Samuel Kounev}.} \bibinfo{year}{2025}\natexlab{}.
\newblock \showarticletitle{WoundAIssist: A Patient-Centered Mobile App for AI-Assisted Wound Care With Physicians in the Loop}.
\newblock \bibinfo{journal}{\emph{arXiv preprint arXiv:2506.06104}} (\bibinfo{year}{2025}).
\newblock


\bibitem[Brown et~al\mbox{.}(2020)]%
        {brown2020language}
\bibfield{author}{\bibinfo{person}{Tom Brown}, \bibinfo{person}{Benjamin Mann}, \bibinfo{person}{Nick Ryder}, \bibinfo{person}{Melanie Subbiah}, \bibinfo{person}{Jared~D Kaplan}, \bibinfo{person}{Prafulla Dhariwal}, \bibinfo{person}{Arvind Neelakantan}, \bibinfo{person}{Pranav Shyam}, \bibinfo{person}{Girish Sastry}, \bibinfo{person}{Amanda Askell}, {et~al\mbox{.}}} \bibinfo{year}{2020}\natexlab{}.
\newblock \showarticletitle{Language models are few-shot learners}.
\newblock \bibinfo{journal}{\emph{Advances in neural information processing systems}}  \bibinfo{volume}{33} (\bibinfo{year}{2020}), \bibinfo{pages}{1877--1901}.
\newblock


\bibitem[Busaranuvong et~al\mbox{.}(2025)]%
        {busaranuvong2025explainable}
\bibfield{author}{\bibinfo{person}{Palawat Busaranuvong}, \bibinfo{person}{Emmanuel Agu}, \bibinfo{person}{Reza~Saadati Fard}, \bibinfo{person}{Deepak Kumar}, \bibinfo{person}{Shefalika Gautam}, \bibinfo{person}{Bengisu Tulu}, {and} \bibinfo{person}{Diane Strong}.} \bibinfo{year}{2025}\natexlab{}.
\newblock \showarticletitle{Explainable, Multi-modal Wound Infection Classification from Images Augmented with Generated Captions}.
\newblock \bibinfo{journal}{\emph{arXiv preprint arXiv:2502.20277}} (\bibinfo{year}{2025}).
\newblock


\bibitem[Cassidy et~al\mbox{.}(2021)]%
        {cassidy2021dfuc}
\bibfield{author}{\bibinfo{person}{Bill Cassidy}, \bibinfo{person}{Neil~D Reeves}, \bibinfo{person}{Joseph~M Pappachan}, \bibinfo{person}{David Gillespie}, \bibinfo{person}{Claire O’Shea}, \bibinfo{person}{Satyan Rajbhandari}, \bibinfo{person}{Arun~G Maiya}, \bibinfo{person}{Eibe Frank}, \bibinfo{person}{Andrew~JM Boulton}, \bibinfo{person}{David~G Armstrong}, {et~al\mbox{.}}} \bibinfo{year}{2021}\natexlab{}.
\newblock \showarticletitle{The DFUC 2020 dataset: analysis towards diabetic foot ulcer detection}.
\newblock \bibinfo{journal}{\emph{touchREVIEWS in Endocrinology}} \bibinfo{volume}{17}, \bibinfo{number}{1} (\bibinfo{year}{2021}), \bibinfo{pages}{5}.
\newblock


\bibitem[Chen et~al\mbox{.}(2024)]%
        {chen2024huatuogpt}
\bibfield{author}{\bibinfo{person}{Junying Chen}, \bibinfo{person}{Chi Gui}, \bibinfo{person}{Ruyi Ouyang}, \bibinfo{person}{Anningzhe Gao}, \bibinfo{person}{Shunian Chen}, \bibinfo{person}{Guiming~Hardy Chen}, \bibinfo{person}{Xidong Wang}, \bibinfo{person}{Ruifei Zhang}, \bibinfo{person}{Zhenyang Cai}, \bibinfo{person}{Ke Ji}, {et~al\mbox{.}}} \bibinfo{year}{2024}\natexlab{}.
\newblock \showarticletitle{Huatuogpt-vision, towards injecting medical visual knowledge into multimodal llms at scale}.
\newblock \bibinfo{journal}{\emph{arXiv preprint arXiv:2406.19280}} (\bibinfo{year}{2024}).
\newblock


\bibitem[Chen and Hong(2024)]%
        {chen2024medblip}
\bibfield{author}{\bibinfo{person}{Qiuhui Chen} {and} \bibinfo{person}{Yi Hong}.} \bibinfo{year}{2024}\natexlab{}.
\newblock \showarticletitle{Medblip: Bootstrapping language-image pre-training from 3d medical images and texts}. In \bibinfo{booktitle}{\emph{Proceedings of the Asian conference on computer vision}}. \bibinfo{pages}{2404--2420}.
\newblock


\bibitem[Chunlin et~al\mbox{.}(2023)]%
        {chunlin2023deep}
\bibfield{author}{\bibinfo{person}{ZHAO Chunlin}, \bibinfo{person}{HE Tingting}, \bibinfo{person}{YUAN Linyan}, \bibinfo{person}{YANG Xue}, \bibinfo{person}{WANG Jing}, \bibinfo{person}{CHEN Xiao}, \bibinfo{person}{LIANG Zhimin}, \bibinfo{person}{GUO Yuchen}, {et~al\mbox{.}}} \bibinfo{year}{2023}\natexlab{}.
\newblock \showarticletitle{Deep learning-based identification of common complication features of surgical incisions}.
\newblock \bibinfo{journal}{\emph{Journal of Sichuan University (Medical Science Edition)}} \bibinfo{volume}{54}, \bibinfo{number}{5} (\bibinfo{year}{2023}).
\newblock


\bibitem[{Craus‑Miguel et al.}(2024)]%
        {RedScar2024_App}
\bibfield{author}{\bibinfo{person}{{Craus‑Miguel et al.}}} \bibinfo{year}{2024}\natexlab{}.
\newblock \bibinfo{title}{RedScar© – Surgical Wound Monitoring (Android App)}.
\newblock \bibinfo{howpublished}{\url{https://play.google.com/store/apps/details?id=com.marcmunar.redscar}}.
\newblock
\newblock
\shownote{Accessed: 2025‑07‑21; AI‑powered post‑operative wound analysis with automated SSI risk scoring}.


\bibitem[Fletcher et~al\mbox{.}(2021a)]%
        {fletcher2021use}
\bibfield{author}{\bibinfo{person}{Richard~Rib{\'o}n Fletcher}, \bibinfo{person}{Gabriel Schneider}, \bibinfo{person}{Laban Bikorimana}, \bibinfo{person}{Gilbert Rukundo}, \bibinfo{person}{Anne Niyigena}, \bibinfo{person}{Elizabeth Miranda}, \bibinfo{person}{Robert Riviello}, \bibinfo{person}{Fredrick Kateera}, {and} \bibinfo{person}{Bethany Hedt-Gauthier}.} \bibinfo{year}{2021}\natexlab{a}.
\newblock \showarticletitle{The use of mobile thermal imaging and deep learning for prediction of surgical site infection}. In \bibinfo{booktitle}{\emph{2021 43rd Annual International Conference of the IEEE Engineering in Medicine \& Biology Society (EMBC)}}. IEEE, \bibinfo{pages}{5059--5062}.
\newblock


\bibitem[Fletcher et~al\mbox{.}(2021b)]%
        {fletcher2021use2}
\bibfield{author}{\bibinfo{person}{Richard~Rib{\'o}n Fletcher}, \bibinfo{person}{Gabriel Schneider}, \bibinfo{person}{Bethany Hedt-Gauthier}, \bibinfo{person}{Theoneste Nkurunziza}, \bibinfo{person}{Barnabas Alayande}, \bibinfo{person}{Robert Riviello}, {and} \bibinfo{person}{Fredrick Kateera}.} \bibinfo{year}{2021}\natexlab{b}.
\newblock \showarticletitle{Use of convolutional neural nets and transfer learning for prediction of surgical site infection from color images}. In \bibinfo{booktitle}{\emph{2021 43rd Annual International Conference of the IEEE Engineering in Medicine \& Biology Society (EMBC)}}. IEEE, \bibinfo{pages}{5047--5050}.
\newblock


\bibitem[He et~al\mbox{.}(2016)]%
        {he2016deep}
\bibfield{author}{\bibinfo{person}{Kaiming He}, \bibinfo{person}{Xiangyu Zhang}, \bibinfo{person}{Shaoqing Ren}, {and} \bibinfo{person}{Jian Sun}.} \bibinfo{year}{2016}\natexlab{}.
\newblock \showarticletitle{Deep residual learning for image recognition}. In \bibinfo{booktitle}{\emph{Proceedings of the IEEE conference on computer vision and pattern recognition}}. \bibinfo{pages}{770--778}.
\newblock


\bibitem[{Healthy.io Ltd.}(2024)]%
        {HealthyIO2024_Minuteful}
\bibfield{author}{\bibinfo{person}{{Healthy.io Ltd.}}} \bibinfo{year}{2024}\natexlab{}.
\newblock \bibinfo{title}{Minuteful for Wound (Android App)}.
\newblock \bibinfo{howpublished}{\url{https://play.google.com/store/apps/details?id=healthy.wound.patient}}.
\newblock
\newblock
\shownote{Accessed: 2025‑07‑21; AI‑powered wound dimension measurement, segmentation, and secure clinician integration}.


\bibitem[Hu et~al\mbox{.}(2022)]%
        {hu2022lora}
\bibfield{author}{\bibinfo{person}{Edward~J Hu}, \bibinfo{person}{Yelong Shen}, \bibinfo{person}{Phillip Wallis}, \bibinfo{person}{Zeyuan Allen-Zhu}, \bibinfo{person}{Yuanzhi Li}, \bibinfo{person}{Shean Wang}, \bibinfo{person}{Lu Wang}, \bibinfo{person}{Weizhu Chen}, {et~al\mbox{.}}} \bibinfo{year}{2022}\natexlab{}.
\newblock \showarticletitle{Lora: Low-rank adaptation of large language models.}
\newblock \bibinfo{journal}{\emph{ICLR}} \bibinfo{volume}{1}, \bibinfo{number}{2} (\bibinfo{year}{2022}), \bibinfo{pages}{3}.
\newblock


\bibitem[Iskandar et~al\mbox{.}(2019)]%
        {iskandar2019highlighting}
\bibfield{author}{\bibinfo{person}{Katia Iskandar}, \bibinfo{person}{Massimo Sartelli}, \bibinfo{person}{Marwan Tabbal}, \bibinfo{person}{Luca Ansaloni}, \bibinfo{person}{Gian~Luca Baiocchi}, \bibinfo{person}{Fausto Catena}, \bibinfo{person}{Federico Coccolini}, \bibinfo{person}{Mainul Haque}, \bibinfo{person}{Francesco~Maria Labricciosa}, \bibinfo{person}{Ayad Moghabghab}, {et~al\mbox{.}}} \bibinfo{year}{2019}\natexlab{}.
\newblock \showarticletitle{Highlighting the gaps in quantifying the economic burden of surgical site infections associated with antimicrobial-resistant bacteria}.
\newblock \bibinfo{journal}{\emph{World Journal of Emergency Surgery}} \bibinfo{volume}{14}, \bibinfo{number}{1} (\bibinfo{year}{2019}), \bibinfo{pages}{50}.
\newblock


\bibitem[Li et~al\mbox{.}(2023)]%
        {li2023llava}
\bibfield{author}{\bibinfo{person}{Chunyuan Li}, \bibinfo{person}{Cliff Wong}, \bibinfo{person}{Sheng Zhang}, \bibinfo{person}{Naoto Usuyama}, \bibinfo{person}{Haotian Liu}, \bibinfo{person}{Jianwei Yang}, \bibinfo{person}{Tristan Naumann}, \bibinfo{person}{Hoifung Poon}, {and} \bibinfo{person}{Jianfeng Gao}.} \bibinfo{year}{2023}\natexlab{}.
\newblock \showarticletitle{Llava-med: Training a large language-and-vision assistant for biomedicine in one day}.
\newblock \bibinfo{journal}{\emph{Advances in Neural Information Processing Systems}}  \bibinfo{volume}{36} (\bibinfo{year}{2023}), \bibinfo{pages}{28541--28564}.
\newblock


\bibitem[Li et~al\mbox{.}(2022)]%
        {li2022blip}
\bibfield{author}{\bibinfo{person}{Junnan Li}, \bibinfo{person}{Dongxu Li}, \bibinfo{person}{Caiming Xiong}, {and} \bibinfo{person}{Steven Hoi}.} \bibinfo{year}{2022}\natexlab{}.
\newblock \showarticletitle{Blip: Bootstrapping language-image pre-training for unified vision-language understanding and generation}. In \bibinfo{booktitle}{\emph{International conference on machine learning}}. PMLR, \bibinfo{pages}{12888--12900}.
\newblock


\bibitem[Lin(2004)]%
        {lin2004rouge}
\bibfield{author}{\bibinfo{person}{Chin-Yew Lin}.} \bibinfo{year}{2004}\natexlab{}.
\newblock \showarticletitle{Rouge: A package for automatic evaluation of summaries}. In \bibinfo{booktitle}{\emph{Text summarization branches out}}. \bibinfo{pages}{74--81}.
\newblock


\bibitem[Liu et~al\mbox{.}(2023)]%
        {liu2023visual}
\bibfield{author}{\bibinfo{person}{Haotian Liu}, \bibinfo{person}{Chunyuan Li}, \bibinfo{person}{Qingyang Wu}, {and} \bibinfo{person}{Yong~Jae Lee}.} \bibinfo{year}{2023}\natexlab{}.
\newblock \showarticletitle{Visual instruction tuning}.
\newblock \bibinfo{journal}{\emph{Advances in neural information processing systems}}  \bibinfo{volume}{36} (\bibinfo{year}{2023}), \bibinfo{pages}{34892--34916}.
\newblock


\bibitem[McLean et~al\mbox{.}(2023a)]%
        {mclean2023readiness}
\bibfield{author}{\bibinfo{person}{Kenneth~A McLean}, \bibinfo{person}{Stephen~R Knight}, \bibinfo{person}{Thomas~M Diehl}, \bibinfo{person}{Chris Varghese}, \bibinfo{person}{Nathan Ng}, \bibinfo{person}{Mark~A Potter}, \bibinfo{person}{Syed~Nabeel Zafar}, \bibinfo{person}{Matt-Mouley Bouamrane}, {and} \bibinfo{person}{Ewen~M Harrison}.} \bibinfo{year}{2023}\natexlab{a}.
\newblock \showarticletitle{Readiness for implementation of novel digital health interventions for postoperative monitoring: a systematic review and clinical innovation network analysis}.
\newblock \bibinfo{journal}{\emph{The Lancet Digital Health}} \bibinfo{volume}{5}, \bibinfo{number}{5} (\bibinfo{year}{2023}), \bibinfo{pages}{e295--e315}.
\newblock


\bibitem[McLean et~al\mbox{.}(2021)]%
        {mclean2021remote}
\bibfield{author}{\bibinfo{person}{Kenneth~A McLean}, \bibinfo{person}{Katie~E Mountain}, \bibinfo{person}{Catherine~A Shaw}, \bibinfo{person}{Thomas~M Drake}, \bibinfo{person}{Riinu Pius}, \bibinfo{person}{Stephen~R Knight}, \bibinfo{person}{Cameron~J Fairfield}, \bibinfo{person}{Alessandro Sgr{\`o}}, \bibinfo{person}{Matt Bouamrane}, \bibinfo{person}{William~A Cambridge}, {et~al\mbox{.}}} \bibinfo{year}{2021}\natexlab{}.
\newblock \showarticletitle{Remote diagnosis of surgical-site infection using a mobile digital intervention: a randomised controlled trial in emergency surgery patients}.
\newblock \bibinfo{journal}{\emph{NPJ Digital Medicine}} \bibinfo{volume}{4}, \bibinfo{number}{1} (\bibinfo{year}{2021}), \bibinfo{pages}{160}.
\newblock


\bibitem[McLean et~al\mbox{.}(2023b)]%
        {mclean2023evaluation}
\bibfield{author}{\bibinfo{person}{Kenneth~A McLean}, \bibinfo{person}{Alessandro Sgr{\`o}}, \bibinfo{person}{Leo~R Brown}, \bibinfo{person}{Louis~F Buijs}, \bibinfo{person}{Luke Daines}, \bibinfo{person}{Mark~A Potter}, \bibinfo{person}{Matt-Mouley Bouamrane}, {and} \bibinfo{person}{Ewen~M Harrison}.} \bibinfo{year}{2023}\natexlab{b}.
\newblock \showarticletitle{Evaluation of remote digital postoperative wound monitoring in routine surgical practice}.
\newblock \bibinfo{journal}{\emph{npj Digital Medicine}} \bibinfo{volume}{6}, \bibinfo{number}{1} (\bibinfo{year}{2023}), \bibinfo{pages}{85}.
\newblock


\bibitem[McLean et~al\mbox{.}(2025)]%
        {mclean2025multimodal}
\bibfield{author}{\bibinfo{person}{Kenneth~A McLean}, \bibinfo{person}{Alessandro Sgr{\`o}}, \bibinfo{person}{Leo~R Brown}, \bibinfo{person}{Louis~F Buijs}, \bibinfo{person}{Katie~E Mountain}, \bibinfo{person}{Catherine~A Shaw}, \bibinfo{person}{Thomas~M Drake}, \bibinfo{person}{Riinu Pius}, \bibinfo{person}{Stephen~R Knight}, \bibinfo{person}{Cameron~J Fairfield}, {et~al\mbox{.}}} \bibinfo{year}{2025}\natexlab{}.
\newblock \showarticletitle{Multimodal machine learning to predict surgical site infection with healthcare workload impact assessment}.
\newblock \bibinfo{journal}{\emph{npj Digital Medicine}} \bibinfo{volume}{8}, \bibinfo{number}{1} (\bibinfo{year}{2025}), \bibinfo{pages}{121}.
\newblock


\bibitem[{Medline Industries, Inc.}(2025)]%
        {Medline2025_SkinHealth}
\bibfield{author}{\bibinfo{person}{{Medline Industries, Inc.}}} \bibinfo{year}{2025}\natexlab{}.
\newblock \bibinfo{title}{Medline Skin Health (iOS App)}.
\newblock \bibinfo{howpublished}{\url{https://apps.apple.com/us/app/medline-skin-health/id1583099820}}.
\newblock
\newblock
\shownote{Accessed: 2025‑07‑31; free app; skin‑ and wound‑care reference library, product selector, formulary management, trending best practices, NE1 wound‑measurement demo}.


\bibitem[Moor et~al\mbox{.}(2023)]%
        {moor2023med}
\bibfield{author}{\bibinfo{person}{Michael Moor}, \bibinfo{person}{Qian Huang}, \bibinfo{person}{Shirley Wu}, \bibinfo{person}{Michihiro Yasunaga}, \bibinfo{person}{Yash Dalmia}, \bibinfo{person}{Jure Leskovec}, \bibinfo{person}{Cyril Zakka}, \bibinfo{person}{Eduardo~Pontes Reis}, {and} \bibinfo{person}{Pranav Rajpurkar}.} \bibinfo{year}{2023}\natexlab{}.
\newblock \showarticletitle{Med-flamingo: a multimodal medical few-shot learner}. In \bibinfo{booktitle}{\emph{Machine Learning for Health (ML4H)}}. PMLR, \bibinfo{pages}{353--367}.
\newblock


\bibitem[Ouyang et~al\mbox{.}(2022)]%
        {ouyang2022training}
\bibfield{author}{\bibinfo{person}{Long Ouyang}, \bibinfo{person}{Jeffrey Wu}, \bibinfo{person}{Xu Jiang}, \bibinfo{person}{Diogo Almeida}, \bibinfo{person}{Carroll Wainwright}, \bibinfo{person}{Pamela Mishkin}, \bibinfo{person}{Chong Zhang}, \bibinfo{person}{Sandhini Agarwal}, \bibinfo{person}{Katarina Slama}, \bibinfo{person}{Alex Ray}, {et~al\mbox{.}}} \bibinfo{year}{2022}\natexlab{}.
\newblock \showarticletitle{Training language models to follow instructions with human feedback}.
\newblock \bibinfo{journal}{\emph{Advances in neural information processing systems}}  \bibinfo{volume}{35} (\bibinfo{year}{2022}), \bibinfo{pages}{27730--27744}.
\newblock


\bibitem[Owens and Stoessel(2008)]%
        {owens2008surgical}
\bibfield{author}{\bibinfo{person}{CD Owens} {and} \bibinfo{person}{K Stoessel}.} \bibinfo{year}{2008}\natexlab{}.
\newblock \showarticletitle{Surgical site infections: epidemiology, microbiology and prevention}.
\newblock \bibinfo{journal}{\emph{Journal of hospital infection}}  \bibinfo{volume}{70} (\bibinfo{year}{2008}), \bibinfo{pages}{3--10}.
\newblock


\bibitem[Papineni et~al\mbox{.}(2002)]%
        {papineni2002bleu}
\bibfield{author}{\bibinfo{person}{Kishore Papineni}, \bibinfo{person}{Salim Roukos}, \bibinfo{person}{Todd Ward}, {and} \bibinfo{person}{Wei-Jing Zhu}.} \bibinfo{year}{2002}\natexlab{}.
\newblock \showarticletitle{Bleu: a method for automatic evaluation of machine translation}. In \bibinfo{booktitle}{\emph{Proceedings of the 40th annual meeting of the Association for Computational Linguistics}}. \bibinfo{pages}{311--318}.
\newblock


\bibitem[{PointClickCare}(2025)]%
        {PointClickCare2025_SkinAndWound}
\bibfield{author}{\bibinfo{person}{{PointClickCare}}.} \bibinfo{year}{2025}\natexlab{}.
\newblock \bibinfo{title}{PointClickCare Skin and Wound (iOS App)}.
\newblock \bibinfo{howpublished}{\url{https://apps.apple.com/us/app/pointclickcare-skin-and-wound/id1073855679}}.
\newblock
\newblock
\shownote{Accessed: 2025‑07‑31; automated wound assessment, non‑contact wound measurements, image capture, graphical trending, auto‑population of MDS Section M, standardized bedside documentation, EHR integration}.


\bibitem[Rochon et~al\mbox{.}(2023)]%
        {rochon2023image}
\bibfield{author}{\bibinfo{person}{M Rochon}, \bibinfo{person}{A Jawarchan}, \bibinfo{person}{F Fagan}, \bibinfo{person}{JA Otter}, {and} \bibinfo{person}{J Tanner}.} \bibinfo{year}{2023}\natexlab{}.
\newblock \showarticletitle{Image-based digital post-discharge surveillance in England: measuring patient enrolment, engagement, clinician response times, surgical site infection, and carbon footprint}.
\newblock \bibinfo{journal}{\emph{Journal of Hospital Infection}}  \bibinfo{volume}{133} (\bibinfo{year}{2023}), \bibinfo{pages}{15--22}.
\newblock


\bibitem[Rochon et~al\mbox{.}(2025)]%
        {rochon2025remote}
\bibfield{author}{\bibinfo{person}{Melissa Rochon}, \bibinfo{person}{Kylie Sandy-Hodgetts}, \bibinfo{person}{Ria Betteridge}, \bibinfo{person}{James Glasbey}, \bibinfo{person}{Kumbi Kariwo}, \bibinfo{person}{Kenneth McLean}, \bibinfo{person}{Jeffrey~A Niezgoda}, \bibinfo{person}{Thomas Serena}, \bibinfo{person}{William~H Tettelbach}, \bibinfo{person}{George Smith}, {et~al\mbox{.}}} \bibinfo{year}{2025}\natexlab{}.
\newblock \showarticletitle{Remote digital surgical wound monitoring and surveillance using smartphones}.
\newblock \bibinfo{journal}{\emph{Journal of Wound Care}} \bibinfo{volume}{34}, \bibinfo{number}{Sup4b} (\bibinfo{year}{2025}), \bibinfo{pages}{S1--S25}.
\newblock


\bibitem[Rochon et~al\mbox{.}(2024a)]%
        {rochon2024post}
\bibfield{author}{\bibinfo{person}{Melissa Rochon}, \bibinfo{person}{Judith Tanner}, \bibinfo{person}{Karen Cariaga}, \bibinfo{person}{Sean~Derick Ingusan}, \bibinfo{person}{Angila Jawarchan}, \bibinfo{person}{Carlos Morais}, \bibinfo{person}{Bella Odattil}, {and} \bibinfo{person}{Ron Dizon}.} \bibinfo{year}{2024}\natexlab{a}.
\newblock \showarticletitle{Post-discharge surgical site infection surveillance using patient smartphones: a single-centre experience in cardiac surgery}.
\newblock \bibinfo{journal}{\emph{British Journal of Healthcare Management}} \bibinfo{volume}{30}, \bibinfo{number}{7} (\bibinfo{year}{2024}), \bibinfo{pages}{1--11}.
\newblock


\bibitem[Rochon et~al\mbox{.}(2024b)]%
        {rochon2024wound}
\bibfield{author}{\bibinfo{person}{Melissa Rochon}, \bibinfo{person}{Judith Tanner}, \bibinfo{person}{James Jurkiewicz}, \bibinfo{person}{Jacqueline Beckhelling}, \bibinfo{person}{Akuha Aondoakaa}, \bibinfo{person}{Keith Wilson}, \bibinfo{person}{Luxmi Dhoonmoon}, \bibinfo{person}{Max Underwood}, \bibinfo{person}{Lara Mason}, \bibinfo{person}{Roy Harris}, {et~al\mbox{.}}} \bibinfo{year}{2024}\natexlab{b}.
\newblock \showarticletitle{Wound imaging software and digital platform to assist review of surgical wounds using patient smartphones: The development and evaluation of artificial intelligence (WISDOM AI study)}.
\newblock \bibinfo{journal}{\emph{PloS one}} \bibinfo{volume}{19}, \bibinfo{number}{12} (\bibinfo{year}{2024}), \bibinfo{pages}{e0315384}.
\newblock


\bibitem[Sandy-Hodgetts et~al\mbox{.}(2022)]%
        {sandy2022non}
\bibfield{author}{\bibinfo{person}{Kylie Sandy-Hodgetts}, \bibinfo{person}{Richard Norman}, \bibinfo{person}{Stephen Edmondston}, \bibinfo{person}{Zaheerah Haywood}, \bibinfo{person}{Leigh Davies}, \bibinfo{person}{Katrina Hulsdunk}, \bibinfo{person}{Jessica Barlow}, {and} \bibinfo{person}{Piers Yates}.} \bibinfo{year}{2022}\natexlab{}.
\newblock \showarticletitle{A non-randomised pragmatic trial for the early detection and prevention of surgical wound complications using an advanced hydropolymer wound dressing and smartphone technology: the EDISON trial protocol}.
\newblock \bibinfo{journal}{\emph{International Wound Journal}} \bibinfo{volume}{19}, \bibinfo{number}{8} (\bibinfo{year}{2022}), \bibinfo{pages}{2174--2182}.
\newblock


\bibitem[Shenoy et~al\mbox{.}(2018)]%
        {shenoy2018deepwound}
\bibfield{author}{\bibinfo{person}{Varun~N Shenoy}, \bibinfo{person}{Elizabeth Foster}, \bibinfo{person}{Lauren Aalami}, \bibinfo{person}{Bakar Majeed}, {and} \bibinfo{person}{Oliver Aalami}.} \bibinfo{year}{2018}\natexlab{}.
\newblock \showarticletitle{Deepwound: Automated postoperative wound assessment and surgical site surveillance through convolutional neural networks}. In \bibinfo{booktitle}{\emph{2018 IEEE International Conference on Bioinformatics and Biomedicine (BIBM)}}. IEEE, \bibinfo{pages}{1017--1021}.
\newblock


\bibitem[Simonyan and Zisserman(2014)]%
        {simonyan2014very}
\bibfield{author}{\bibinfo{person}{Karen Simonyan} {and} \bibinfo{person}{Andrew Zisserman}.} \bibinfo{year}{2014}\natexlab{}.
\newblock \showarticletitle{Very deep convolutional networks for large-scale image recognition}.
\newblock \bibinfo{journal}{\emph{arXiv preprint arXiv:1409.1556}} (\bibinfo{year}{2014}).
\newblock


\bibitem[{Synergy Wound Technology, LLC}(2025)]%
        {Synergy2025_InteliWound}
\bibfield{author}{\bibinfo{person}{{Synergy Wound Technology, LLC}}.} \bibinfo{year}{2025}\natexlab{}.
\newblock \bibinfo{title}{InteliWound (iOS App)}.
\newblock \bibinfo{howpublished}{\url{https://apps.apple.com/us/app/inteliwound/id1313983915}}.
\newblock
\newblock
\shownote{Accessed: 2025‑07‑31; value‑based wound management platform—guided assessments, volumetric wound measurements, treatment suggestions, product instructions, scoring, supply recommendations, standardized documentation, clinician collaboration}.


\bibitem[Tabja~Bortesi et~al\mbox{.}(2024)]%
        {tabja2024machine}
\bibfield{author}{\bibinfo{person}{Juan~Pablo Tabja~Bortesi}, \bibinfo{person}{Jonathan Ranisau}, \bibinfo{person}{Shuang Di}, \bibinfo{person}{Michael McGillion}, \bibinfo{person}{Laura Rosella}, \bibinfo{person}{Alistair Johnson}, \bibinfo{person}{PJ Devereaux}, {and} \bibinfo{person}{Jeremy Petch}.} \bibinfo{year}{2024}\natexlab{}.
\newblock \showarticletitle{Machine learning approaches for the image-based identification of surgical wound infections: scoping review}.
\newblock \bibinfo{journal}{\emph{Journal of Medical Internet Research}}  \bibinfo{volume}{26} (\bibinfo{year}{2024}), \bibinfo{pages}{e52880}.
\newblock


\bibitem[Wang et~al\mbox{.}(2024)]%
        {wang2024qwen2}
\bibfield{author}{\bibinfo{person}{Peng Wang}, \bibinfo{person}{Shuai Bai}, \bibinfo{person}{Sinan Tan}, \bibinfo{person}{Shijie Wang}, \bibinfo{person}{Zhihao Fan}, \bibinfo{person}{Jinze Bai}, \bibinfo{person}{Keqin Chen}, \bibinfo{person}{Xuejing Liu}, \bibinfo{person}{Jialin Wang}, \bibinfo{person}{Wenbin Ge}, {et~al\mbox{.}}} \bibinfo{year}{2024}\natexlab{}.
\newblock \showarticletitle{Qwen2-vl: Enhancing vision-language model's perception of the world at any resolution}.
\newblock \bibinfo{journal}{\emph{arXiv preprint arXiv:2409.12191}} (\bibinfo{year}{2024}).
\newblock


\bibitem[Wu et~al\mbox{.}(2020)]%
        {wu2020unified}
\bibfield{author}{\bibinfo{person}{Jin-Ming Wu}, \bibinfo{person}{Chia-Jui Tsai}, \bibinfo{person}{Te-Wei Ho}, \bibinfo{person}{Feipei Lai}, \bibinfo{person}{Hao-Chih Tai}, {and} \bibinfo{person}{Ming-Tsan Lin}.} \bibinfo{year}{2020}\natexlab{}.
\newblock \showarticletitle{A unified framework for automatic detection of wound infection with artificial intelligence}.
\newblock \bibinfo{journal}{\emph{Applied Sciences}} \bibinfo{volume}{10}, \bibinfo{number}{15} (\bibinfo{year}{2020}), \bibinfo{pages}{5353}.
\newblock


\bibitem[Zhang et~al\mbox{.}(2019)]%
        {zhang2019bertscore}
\bibfield{author}{\bibinfo{person}{Tianyi Zhang}, \bibinfo{person}{Varsha Kishore}, \bibinfo{person}{Felix Wu}, \bibinfo{person}{Kilian~Q Weinberger}, {and} \bibinfo{person}{Yoav Artzi}.} \bibinfo{year}{2019}\natexlab{}.
\newblock \showarticletitle{Bertscore: Evaluating text generation with bert}.
\newblock \bibinfo{journal}{\emph{arXiv preprint arXiv:1904.09675}} (\bibinfo{year}{2019}).
\newblock


\end{thebibliography}

\appendix

\section{Dataset Information}

\subsection{Dataset Construction}

\noindent \textbf{Data Collection} We collect surgical wound images from publicly available content on social media platforms. Specifically, we utilize a collection of domain-specific hashtags and keywords to extract relevant content from various platforms, including \textit{RedNote}, \textit{Twitter}, \textit{Facebook}, \textit{Instagram}, and \textit{Reddit}. In addition, we further expand the dataset by collecting images specifically from the social media accounts of surgeons and other medical professionals, where postoperative wound cases are often shared for educational or awareness purposes. Details of the hashtags, keywords, and surgeons' accounts used are shown in Table 7.

\begin{table}[h]
  \centering
  \caption{Social Media Metadata for Surgical Wound Research}
  \label{tab:hashtags_keywords_surgeons}
  \begin{tabular}{|l|l|}
    \hline
    \textbf{Category} & \textbf{Content} \\
    \hline
    Hashtags & \makecell{\#surgicalwound,\\  \#surgicalwounds,\\ \#woundinfection, \\ (\#surgical OR \#wound),\\ (\#postoperative OR \#wound), \\ \#woundtreatment} \\
    \hline
    Keywords & \makecell{surgical wound,\\ surgical wound infection,\\ surgical site infection, \\ surgical wound erythema,\\ surgical wound edema, \\ surgical wound suppuration,\\ postoperative wound, \\ postoperative wound infection,\\ postoperative wound site infection, \\ postoperative wound erythema,\\ postoperative wound edema, \\ postoperative wound suppuration} \\
    \hline
    Surgeons’ Accounts & \makecell{@anikin} \\
    \hline
  \end{tabular}
\end{table}

\noindent \textbf{Data Filtering} We employ a two-stage filtering process that involves both AI and human expert reviews to ensure that only high-quality images containing visible surgical wounds are included in the dataset.
First, we leverage GPT-4o as an AI expert to automatically assess whether an image depicts a clear surgical wound, filtering out low-resolution images or those lacking wound-related content.
Subsequently, three surgeons act as human experts to manually review the remaining images and exclude any that are low-resolution or do not depict authentic surgical wounds.
According to the statistics, a total of 71 images were filtered.

\noindent \textbf{Expert Annotation} To ensure high-quality annotation while optimizing expert effort, we estimate the difficulty level of each image based on the predicted risk levels from three MLLMs: GPT-4o, Claude 3.5, and Gemini 2. Images for which all three models consistently predict a low risk level are considered low-difficulty, whereas those with inconsistent predictions or predicted as medium or high risk are categorized as high-difficulty. According to our data, 90 images were classified as high-difficulty. For low-difficulty cases, a single surgeon is randomly assigned to perform the annotation. For high-difficulty cases, the image is independently annotated by two randomly assigned surgeons. If any disagreement arises between their annotations, a third surgeon reviews both sets and makes the final decision.
The platform annotated by doctors is shown in Figure 4. Surgical wound images are displayed on the left, surgical wound features and diagnostic results—annotated through eight multiple-choice questions—are presented in the center, and a report generation window is located on the right. GPT-4o generates corresponding reports based on the images and annotations provided by the surgeons, which are then reviewed and refined by the surgeons.

\begin{figure*}[!t]  
  \centering
  \includegraphics[width=\textwidth]{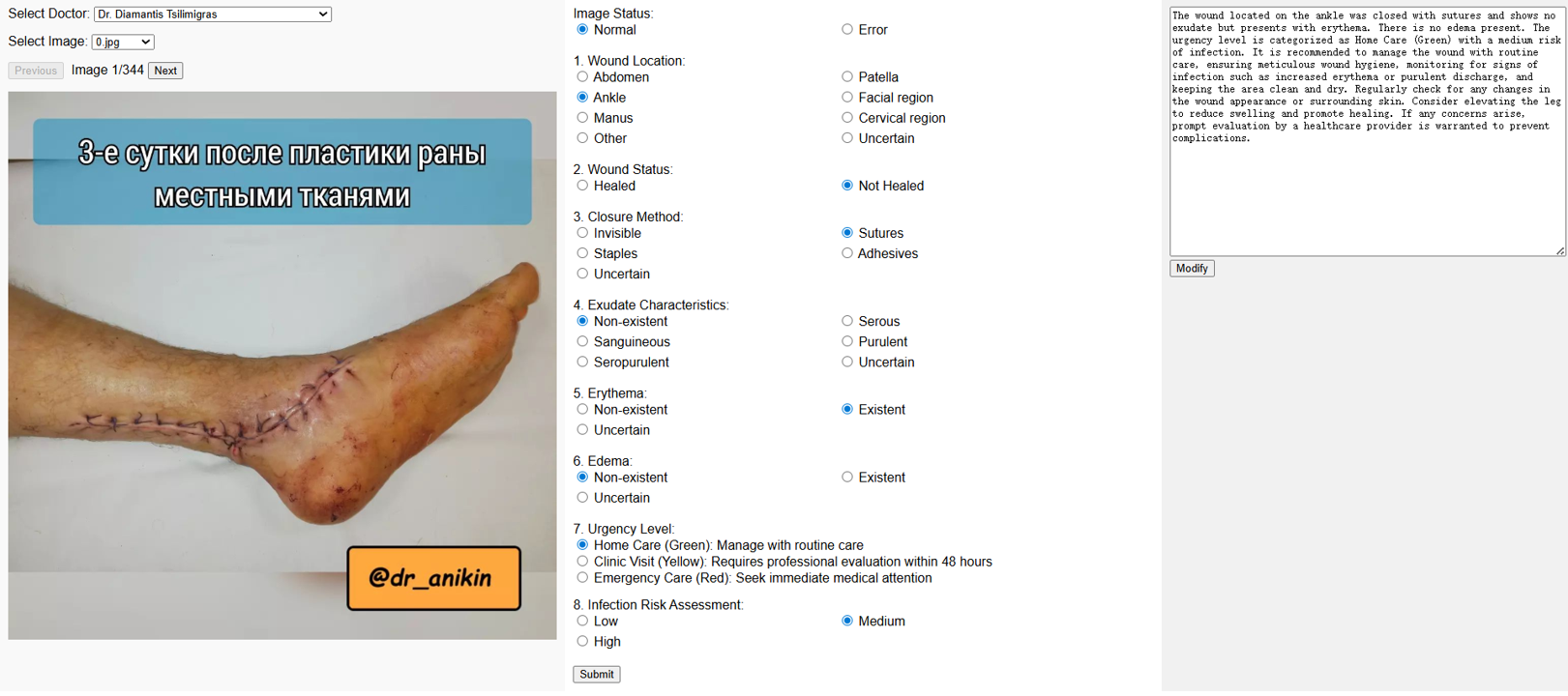}
  \caption{Website for data annotation}
  \Description{}
  \label{}
\end{figure*}

\subsection{Ethics and Fairness Statement}
Data collection and processing meet the website requirements. Additionally, we have de-identified the images, minimizing the risk of privacy loss.

\subsection{Data License}
The dataset is released under the CC BY-SA 4.0 license.

\subsection{Data Access}
Users can explore and download our dataset at 

\noindent https://huggingface.co/datasets/xuxuxuxuxu/SurgWound 
with documentation at https://github.com/xuxuxuxuxuxjh/SurgWound.

\noindent SurgWound dataset is also under submission to PhysioNet.

\subsection{Dataset Analysis}

\noindent \textbf{Independence Analysis Using Chi-Square Test}

\noindent (1) Assumed null hypothesis $H_0$: attribute is independent of infection risk.

\noindent(2) Calculated the $\chi^2$ and p-value:

\begin{equation}
\chi^2 = \sum_{i,j} \frac{(O_{ij} - E_{ij})^2}{E_{ij}}, 
\quad E_{ij} = \frac{O_{i\cdot} \times O_{\cdot j}}{N}
\end{equation}

\begin{equation}
p = P\bigl(\chi^2_{\text{df}} \geq \chi^2 \bigr) = 1 - F_{\chi^2}(\chi^2, \text{df})
\end{equation}

\noindent where $O_{ij}$ is the observed count in cell $(i,j)$, $E_{ij}$ is the expected count under the null hypothesis, $N$ is the total sample size and\(F_{\chi^2}(\cdot)\) is the cumulative distribution function of the chi-square distribution with \(\text{df}=(r-1)(c-1)\).

\noindent (3) Computed Cramér’s V to quantify the strength of these associations:

\begin{equation}
V = \sqrt{\frac{\chi^2}{N \times \min(r-1,\,c-1)}}
\end{equation}

\noindent where $r$ and $c$ are the numbers of rows and columns in the contingency table, respectively, and Cramér’s V ranges between 0 and 1—with higher values indicating stronger association.

\noindent\textbf{Relative Risk Analysis of Surgical Wound Characteristics}

Relative Risk is the ratio of the incidence of outcome (medium or high risk) in the exposed group (e.g., exhibiting a given surgical wound characteristic) versus the non-exposed group. Formally:
\[
\text{RR} = \frac{P(\text{Medium/High Risk} \mid \text{Characteristic present})}{P(\text{Medium/High Risk} \mid \text{Characteristic absent})}
\]
Values of RR interpret as:
\begin{itemize}
  \item \(RR = 1\): No difference in risk compared to baseline,
  \item \(RR > 1\): Elevated risk (potential risk factor),
  \item \(RR < 1\): Reduced risk (potential protective factor)
\end{itemize}

\noindent The detailed experimental results are shown in Figure 5.

\begin{figure*}[!t]  
  \centering
  \includegraphics[width=\textwidth]{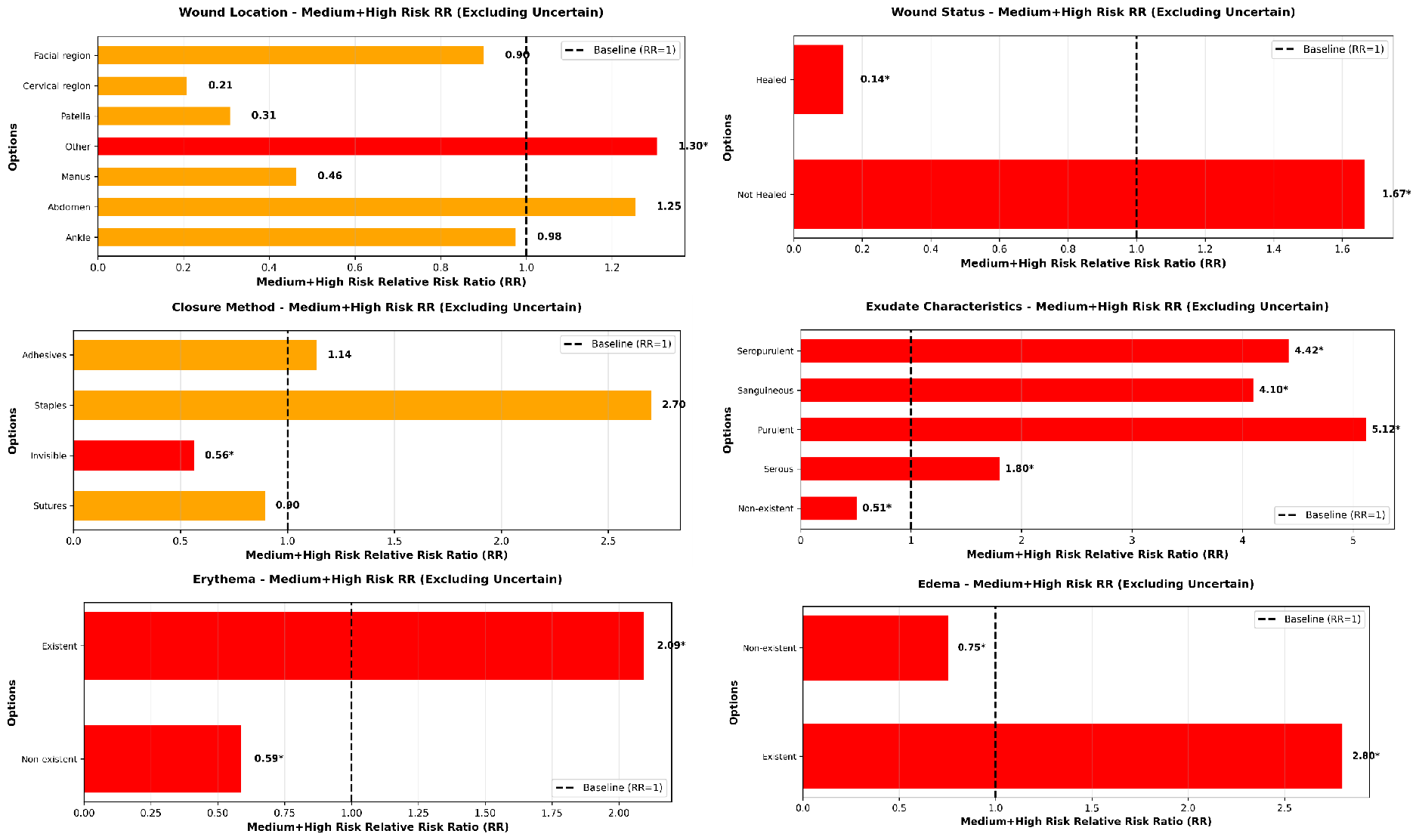}
  \caption{Result in Relative Risk Analysis}
  \Description{}
  \label{}
\end{figure*}

\section{Evaluation Metrics}
\noindent \textbf{VQA Task}
In addition to Accuracy, we compute Precision, Recall, F1-score, Micro-F1, Macro-F1, and Weighted-F1 to better assess model performance under imbalanced conditions. 

\[
\text{Precision} = \frac{\mathrm{TP}}{\mathrm{TP} + \mathrm{FP}}
\]
\[
\text{Recall} = \frac{\mathrm{TP}}{\mathrm{TP} + \mathrm{FN}}, 
\]
\[
F_1 = \frac{2\,\mathrm{Precision}\cdot\mathrm{Recall}}{\mathrm{Precision} + \mathrm{Recall}}
\]

\[
Micro‑F_1 = \frac{2\,\sum_k \mathrm{TP}_k}{2\,\sum_k \mathrm{TP}_k + \sum_k \mathrm{FP}_k + \sum_k \mathrm{FN}_k}
\]

\[
Macro‑F_1 = \frac{1}{K} \sum_{k=1}^K F_{1,k}
\]

\[
Weighted‑F_1 = \frac{1}{\sum_k n_k} \sum_{k=1}^K n_k \cdot F_{1,k}
\]

\noindent where TP denotes the number of true positives, FP the number of false positives, and FN the number of false negatives.

\noindent \textbf{Report Generation Task}
To evaluate report generation performance, we utilize three metrics: BLEU, ROUGE and BERTScore.

\begin{itemize}
  \item \textbf{BLEU} measures the n‑gram precision between generated reports and reference texts, with a brevity penalty to avoid overly short outputs. Scores range from 0 to 1, where higher values indicate closer correspondence to human reference translations.
  \item \textbf{ROUGE} emphasizes n-gram recall (e.g., ROUGE-1, ROUGE-2) and the longest common subsequence (ROUGE-L), which captures the coverage of key content by comparing the generated report against the reference report.
  \item \textbf{BERTScore} BERTScore offers semantic-level evaluation, capturing paraphrase and rephrasing, and providing a higher-fidelity assessment of meaning than surface-based metrics.
\end{itemize}

\section{Experiments}

\subsection{VQA Task}

In addition to Accuracy, we compute Precision, Recall, F1-score, Micro-F1, Macro-F1, and Weighted-F1 to better assess model performance under imbalanced conditions. Tables 8, 9, 10, 11, 12, 13, and 14 respectively present the experimental results of seven VQA sub-tasks: Healing Status, Closure Method, Exudate Type, Erythema, Edema, Urgency Level, and Infection Risk.

\begin{table*}[t]
\centering
\caption{Healing Status VQA task performance}
\begin{tabular}{lcccccc}
\toprule
Model & Healed & Not Healed & Micro‑F1 & Macro‑F1 & Weighted‑F1 \\
      & (P/R/F1) & (P/R/F1) & & & \\
\midrule
\textbf{WoundQwen} & 0.778/0.854/0.814 & 0.745/0.636/0.686 & \textbf{0.766} & \textbf{0.750} & \textbf{0.763} \\
GPT‑4o         & 0.647/0.186/0.289 & 0.593/0.897/0.714 & 0.596 & 0.502 & 0.531 \\
Claude‑3.5     & 0.545/0.102/0.171 & 0.574/0.897/0.700 & 0.563 & 0.436 & 0.472 \\
Gemini‑1.5     & 0.714/0.169/0.274 & 0.602/0.949/0.736 & 0.613 & 0.505 & 0.537 \\
Gemini‑2.0     & 0.619/0.220/0.325 & 0.603/0.897/0.722 & 0.606 & 0.523 & 0.551 \\
HuatuoGPT-34B          & 0.525/0.542/0.533 & 0.629/0.564/0.595 & 0.567 & 0.564 & 0.568 \\
InternVL3-78B      & 0.429/0.051/0.091 & 0.569/0.949/0.711 & 0.562 & 0.401 & 0.444 \\
Qwen2.5‑VL‑72B & 0.609/0.237/0.341 & 0.600/0.846/0.702 & 0.593 & 0.522 & 0.547 \\
\bottomrule
\end{tabular}
\end{table*}

\begin{table*}[h]
\centering
\caption{Closure Method VQA task performance}
\begin{tabular}{lcccccccc}
\toprule
Model & Adhesives & Invisible & Staples & Sutures & Micro‑F1 & Macro‑F1 & Weighted‑F1 \\
      & (P/R/F1) & (P/R/F1) & (P/R/F1) & (P/R/F1) & & & \\
\midrule
\textbf{WoundQwen} & 1.000/0.250/0.400 & 0.632/0.632/0.632 & 0.000/0.000/0.000 & 0.861/0.912/0.886 & \textbf{0.815} & 0.479 & \textbf{0.802} \\
GPT‑4o         & 0.000/0.000/0.000 & 1.000/0.053/0.100 & 1.000/1.000/1.000 & 0.903/0.823/0.861 & 0.744 & 0.490 & 0.668 \\
Claude‑3.5     & 0.000/0.000/0.000 & 0.461/0.316/0.375 & 0.500/1.000/0.667 & 0.873/0.706/0.780 & 0.679 & 0.455 & 0.662 \\
Gemini‑1.5     & 0.000/0.000/0.000 & 1.000/0.105/0.190 & 1.000/1.000/1.000 & 0.898/0.779/0.835 & 0.723 & 0.506 & 0.667 \\
Gemini‑2.0     & 0.000/0.000/0.000 & 0.800/0.210/0.333 & 1.000/1.000/1.000 & 0.853/0.853/0.853 & 0.754 & \textbf{0.546} & 0.710 \\
HuatuoGPT-34B         & 1.000/0.250/0.400 & 0.500/0.158/0.240 & 0.000/0.000/0.000 & 0.791/0.779/0.785 & 0.687 & 0.356 & 0.647 \\
InternVL3-78B      & 0.500/0.250/0.333 & 0.440/0.579/0.500 & 0.200/1.000/0.333 & 0.918/0.662/0.769 & 0.670 & 0.484 & 0.690 \\
Qwen2.5‑VL‑72B & 1.000/0.250/0.400 & 1.000/0.105/0.190 & 0.000/0.000/0.000 & 0.806/0.794/0.800 & 0.704 & 0.348 & 0.648 \\
\bottomrule
\end{tabular}
\end{table*}

\begin{table*}[t]
  \centering
  \caption{Exudate Type VQA task performance}
  \begin{tabular}{p{2cm} p{2cm} p{2cm} p{2cm} p{2cm} p{2cm} p{0.7cm} p{0.7cm} p{0.7cm}}
    \toprule
    Model & Non‑existent & Purulent & Sanguineous & Seropurulent & Serous & Micro & Macro & F1 \\
    \midrule
    \textbf{WoundQwen} & 0.963/0.954/0.959 & 0.000/0.000/0.000 & 0.400/1.000/0.571 & 0.000/0.000/0.000 & 0.600/0.375/0.462 & \textbf{0.888} & \textbf{0.398} & \textbf{0.884} \\
    GPT‑4o                & 1.000/0.376/0.547 & 0.000/0.000/0.000 & 0.200/0.500/0.286 & 0.000/0.000/0.000 & 0.150/0.750/0.250 & 0.437     & 0.216     & 0.502 \\
    Claude‑3.5            & 0.952/0.734/0.829 & 0.000/0.000/0.000 & 1.000/0.500/0.667 & 0.000/0.000/0.000 & 0.154/0.500/0.235 & 0.717     & 0.346     & 0.759 \\
    Gemini‑1.5            & 1.000/0.220/0.361 & 0.000/0.000/0.000 & 0.667/1.000/0.800 & 0.000/0.000/0.000 & 0.062/0.125/0.083 & 0.329     & 0.249     & 0.346 \\
    Gemini‑2.0            & 1.000/0.330/0.497 & 0.000/0.000/0.000 & 0.200/1.000/0.333 & 0.000/0.000/0.000 & 0.000/0.000/0.000 & 0.408     & 0.166     & 0.444 \\
    Huatuo-34B                & 0.969/0.569/0.717 & 0.067/0.333/0.111 & 0.500/0.250/0.333 & 0.000/0.000/0.000 & 0.067/0.125/0.087 & 0.570     & 0.250     & 0.644 \\
    InternVL3-78B             & 1.000/0.541/0.702 & 0.111/0.333/0.167 & 0.167/1.000/0.286 & 0.000/0.000/0.000 & 0.000/0.000/0.000 & 0.556     & 0.231     & 0.626 \\
    Qwen2.5‑VL‑72B        & 0.985/0.605/0.750 & 0.000/0.000/0.000 & 0.174/1.000/0.296 & 0.000/0.000/0.000 & 0.200/0.250/0.222 & 0.595     & 0.254     & 0.678 \\
    \bottomrule
  \end{tabular}
\end{table*}

\begin{table*}[t]
  \centering
  \caption{Erythema VQA task performance}
  \begin{tabular}{lccccc}
    \toprule
    Model & Existent (P/R/F1) & Non‑existent (P/R/F1) & Micro‑F1 & Macro‑F1 & Weighted‑F1 \\
    \midrule
    \textbf{WoundQwen} & 0.523/0.575/0.548 & 0.805/0.769/0.787 & \textbf{0.710} & \textbf{0.667} & \textbf{0.714} \\
    GPT‑4o                & 0.330/0.900/0.483 & 0.895/0.187/0.309 & 0.409     & 0.396     & 0.362       \\
    Claude‑3.5            & 0.336/0.950/0.497 & 0.933/0.154/0.264 & 0.401     & 0.380     & 0.335       \\
    Gemini‑1.5            & 0.330/0.925/0.487 & 1.000/0.088/0.162 & 0.359     & 0.324     & 0.261       \\
    Gemini‑2.0            & 0.358/0.975/0.523 & 1.000/0.198/0.330 & 0.442     & 0.427     & 0.389       \\
    HuatuoGPT-34B                & 0.485/0.400/0.438 & 0.768/0.692/0.728 & 0.642     & 0.583     & 0.554       \\
    InternVL3-78B             & 0.351/0.850/0.496 & 0.818/0.297/0.435 & 0.467     & 0.466     & 0.454       \\
    Qwen2.5‑VL‑72B        & 0.358/0.850/0.504 & 0.824/0.308/0.448 & 0.477     & 0.476     & 0.465       \\
    \bottomrule
  \end{tabular}
\end{table*}

\begin{table*}[t]  
  \centering
  \caption{Erythema VQA task performance}
  \begin{tabular}{lccccc}
    \toprule
    Model & Existent (P/R/F1) & Non‑existent (P/R/F1) & Micro‑F1 & Macro‑F1 & Weighted‑F1 \\
    \midrule
    \textbf{WoundQwen} & 0.393/0.524/0.449 & 0.885/0.819/0.851 & \textbf{0.765} & \textbf{0.650} & \textbf{0.777} \\
    GPT‑4o                & 0.562/0.429/0.486 & 0.875/0.670/0.759 & 0.709     & 0.623     & 0.709       \\
    Claude‑3.5            & 0.261/0.571/0.358 & 0.857/0.574/0.688 & 0.589     & 0.523     & 0.628       \\
    Gemini‑1.5            & 0.250/0.381/0.302 & 0.885/0.245/0.383 & 0.358     & 0.343     & 0.368       \\
    Gemini‑2.0            & 0.375/0.429/0.400 & 0.890/0.691/0.778 & 0.698     & 0.589     & 0.709       \\
    HuatuoGPT-34B                & 0.316/0.571/0.407 & 0.857/0.447/0.587 & 0.535     & 0.497     & 0.554       \\
    InternVL3-78B             & 0.167/0.048/0.074 & 0.828/0.564/0.671 & 0.584     & 0.372     & 0.562       \\
    Qwen2.5‑VL‑72B        & 0.312/0.238/0.270 & 0.832/0.840/0.836 & 0.746     & 0.553     & 0.733       \\
    \bottomrule
  \end{tabular}
  \label{tab:edema}
\end{table*}

\begin{table*}[t]  
  \centering
  \caption{Infection Risk VQA task performance}
  \begin{tabular}{lcccccc}
    \toprule
    Model & Low (P/R/F1) & Medium (P/R/F1) & High (P/R/F1) & Micro‑F1 & Macro‑F1 & Weighted‑F1 \\
    \midrule
    \textbf{WoundQwen} & 0.952/0.877/0.913 & 0.429/0.667/0.522 & 0.500/0.400/0.444 & \textbf{0.832} & \textbf{0.626} & \textbf{0.845} \\
    GPT‑4o & 0.933/0.368/0.528 & 0.159/0.778/0.264 & 0.000/0.000/0.000 & 0.412 & 0.264 & 0.474 \\
    Claude‑3.5 & 0.901/0.561/0.692 & 0.169/0.556/0.260 & 0.000/0.000/0.000 & 0.548 & 0.317 & 0.610 \\
    Gemini‑1.5 & 0.930/0.465/0.620 & 0.147/0.611/0.237 & 0.000/0.000/0.000 & 0.467 & 0.285 & 0.547 \\
    Gemini‑2.0 & 1.000/0.263/0.417 & 0.170/1.000/0.290 & 1.000/0.200/0.333 & 0.358 & 0.347 & 0.397 \\
    HuatuoGPT-34B & 0.930/0.702/0.800 & 0.270/0.556/0.364 & 0.333/0.400/0.364 & 0.692 & 0.509 & 0.727 \\
    InternVL3-78B & 0.943/0.289/0.443 & 0.156/0.778/0.259 & 0.167/0.400/0.235 & 0.358 & 0.312 & 0.411 \\
    Qwen2.5‑VL‑72B & 0.909/0.351/0.506 & 0.141/0.667/0.233 & 0.143/0.200/0.167 & 0.388 & 0.302 & 0.458 \\
    \bottomrule
  \end{tabular}
  \label{tab:infection_risk_assessment}
\end{table*}

\begin{table*}[t]  
  \centering
  \caption{Urgency Level VQA task performance}
  \begin{tabular}{lcccccc}
    \toprule
    Model & Home Care (P/R/F1) & Clinic Visit (P/R/F1) & Emergency Care (P/R/F1) & Micro‑F1 & Macro‑F1 & Weighted‑F1 \\
    \midrule
    \textbf{WoundQwen}       & 0.907/0.907/0.907 & 0.294/0.312/0.303 & 0.000/0.000/0.000 & \textbf{0.832}          & \textbf{0.403}          & \textbf{0.822 }          \\
    GPT‑4o                      & 0.933/0.353/0.512 & 0.135/0.750/0.229 & 0.000/0.000/0.000 & 0.397          & 0.247          & 0.472           \\
    Claude‑3.5                  & 0.941/0.538/0.684 & 0.172/0.688/0.275 & 0.000/0.000/0.000 & 0.556          & 0.320          & 0.627           \\
    Gemini‑1.5                  & 0.957/0.185/0.310 & 0.124/0.875/0.217 & 0.000/0.000/0.000 & 0.263          & 0.176          & 0.294           \\
    Gemini‑2.0                  & 1.000/0.118/0.211 & 0.054/0.125/0.075 & 1.000/0.500/0.667 & 0.180          & 0.318          & 0.201           \\
    HuatuoGPT-34B                      & 0.892/0.832/0.861 & 0.083/0.062/0.071 & 0.333/1.000/0.500 & 0.767          & 0.477          & 0.763           \\
    InternVL3-78B                   & 1.000/0.143/0.250 & 0.107/0.562/0.180 & 0.111/0.500/0.182 & 0.219          & 0.204          & 0.241           \\
    Qwen2.5‑VL‑72B             & 0.930/0.336/0.494 & 0.126/0.687/0.214 & 0.143/0.500/0.222 & 0.380          & 0.310          & 0.457           \\
    \bottomrule
  \end{tabular}
  \label{tab:urgency}
\end{table*}

\subsection{Report Generation Task}

To evaluate report generation performance, we utilize three metrics: BLEU, ROUGE, and BERTScore. Specifically, BLEU includes BLEU-1, BLEU-2, and BLEU-3, while ROUGE comprises ROUGE-1, ROUGE-2, and ROUGE-L. The experimental results are shown in Table 15.

\begin{table*}[h]
  \centering
  \caption{Result in Report Generation Task}
  \begin{tabular}{lccccccc}
    \toprule
    Model & BLEU‑1 & BLEU‑2 & BLEU‑3 & ROUGE‑1 & ROUGE‑2 & ROUGE‑L & BERTScore \\
    \midrule
    \textbf{WoundQwen} & \textbf{0.457} & \textbf{0.300} & \textbf{0.212} & \textbf{0.612} & \textbf{0.279} & \textbf{0.418} & \textbf{0.917} \\
    GPT‑4o          & 0.160  & 0.072  & 0.026  & 0.412   & 0.114   & 0.213   & 0.852 \\
    Claude‑3.5      & 0.170  & 0.073  & 0.026  & 0.391   & 0.097   & 0.205   & 0.854 \\
    Gemini‑1.5      & 0.169  & 0.069  & 0.019  & 0.397   & 0.093   & 0.199   & 0.844 \\
    Gemini‑2.0      & 0.214  & 0.090  & 0.027  & 0.428   & 0.105   & 0.226   & 0.846 \\
    Huatuo-34B          & 0.252 & 0.108 & 0.042 & 0.409 & 0.087 & 0.195 & 0.870 \\
    InternVL3-78B       & 0.153  & 0.053  & 0.011  & 0.367   & 0.078   & 0.192   & 0.837 \\
    Qwen2.5‑VL‑72B  & 0.234  & 0.112  & 0.052  & 0.435   & 0.115   & 0.231   & 0.874 \\
    \bottomrule
  \end{tabular}
\end{table*}

\subsection{APP}
We systematically tested all available remote wound monitoring applications listed on Google Play and the Apple App Store, with the results showing in Table 16.

\begin{table*}[t]
  \centering
  \caption{Evaluation of Wound Monitoring Apps}
  \label{tab:wound-apps}
  \setlength{\tabcolsep}{4.2pt} 
  \begin{tabular}{@{}c c c c c c c c@{}}
    \toprule
    App & Source & Use Domain & Availability &
    \makecell{Internal\\Account} &
    \makecell{Operational\\Status} &
    \makecell{AI Wound\\Measure} &
    \makecell{AI Risk\\Assessment} \\
    \midrule
    Minuteful for Wound & Google Play/App Store & General & × & √ & \longdash & √ & × \\
    WoundTele           & Google Play/App Store & General & × & × & Crash & √ & √ \\
    Tissue Analytics    & Google Play & General & × & √ & \longdash & √ & × \\
    Swift Skin and Wound& Google Play/App Store & General & × & √ & \longdash & √ & × \\
    ECS Wound Tracker   & Google Play & General & × & √ & \longdash & × & × \\
    Wound Wise          & Google Play/App Store & General & × & × & Error & × & × \\
    Wound Compass       & Google Play & General & × & √ & \longdash & × & × \\
    Wound Pros          & Google Play & General & × & √ & \longdash & √ & × \\
    RedScar             & Google Play & Abdominal surgery & × & × & Waiting & √ & × \\
    MyWoundHealing      & Google Play/App Store & General & × & × & Normal & × & × \\
    Medline skin health & App Store   & General & × & × & \longdash & √  & √ \\
    PointClickCare Skin and Wound & App Store & General & × & √ & \longdash & √ & √ \\
    Wound Care Pro      & App Store   & General & √ & × & Normal & × & × \\
    Wound Capture       & App Store   & General & × & × & Connection Fail & √ & × \\
    Archangel WOC Care Platform & App Store   & General & × & √ & \longdash & √ & × \\ 
    Savvy Wound Report  & App Store   & General & × & √ & \longdash & × & × \\
    InteliWound         & App Store   & General & × & √ & \longdash & √ & × \\
    Futura Wound Care   & App Store   & General & × & √ & \longdash & √ & × \\
    WoundSee            & App Store   & General & × & × & Request failed & √ & × \\
   
    \bottomrule
  \end{tabular}
\end{table*}

\end{document}